\documentclass[10pt]{article}
\usepackage[margin=1in]{geometry}

\usepackage{natbib}
\usepackage{hyperref}
\usepackage{url}
\usepackage{placeins}
\usepackage{amsthm}
\usepackage{booktabs}
\usepackage{graphicx}
\usepackage{subcaption}
\usepackage{amsmath, amssymb, amsfonts}
\usepackage{mathtools}
\usepackage{array}
\usepackage{multirow}
\usepackage{tabularx}
\usepackage{threeparttable}
\usepackage{algorithm}
  
\usepackage{algorithmic}
\usepackage{textcomp}
\usepackage{xcolor}
\usepackage{siunitx}
\theoremstyle{plain}

\theoremstyle{definition}

\theoremstyle{remark}

\title{SPORE: Skeleton Propagation Over Recalibrating Expansions}

\author{Randolph Wiredu-Aidoo}

\begin{document}
\maketitle

\begin{abstract}
Many real-world datasets are not linearly separable, limiting the effectiveness of centroid-based clustering methods such as K-means. Density-based clustering methods address this limitation by identifying clusters with arbitrary geometric structure; however, existing approaches exhibit two persistent shortcomings. First, they often underperform in the presence of heterogeneous local densities, where a single density threshold cannot adequately capture clusters across multiple density scales. Second, they generally lack the clear boundary delineation naturally induced by the linear partitioning mechanism of centroid-based methods. This paper introduces SPORE (Skeleton Propagation Over Recalibrating Expansions), a clustering algorithm designed to address both challenges while preserving the geometric flexibility of density-based approaches. SPORE operates in two stages: an adaptive cluster expansion phase followed by a proximity-driven boundary propagation phase that maintains discriminative capability even under weak density contrast. The proposed method is evaluated on 28 benchmark datasets against established density-based baselines, with K-means included as a reference centroid-based method. Experimental results demonstrate that SPORE achieves significantly improved cluster recovery relative to all evaluated baselines ($p < 0.01$), while strong-performing configurations can be identified within five random-search evaluations.
\end{abstract}

\section{Introduction}
\label{sec:intro}
Clustering methods that assume linear separability, with K-means \citep{MacQueen1967} as a canonical example, partition data into convex and approximately isotropic regions. Their boundaries are well-defined by construction, as assignments are determined directly through centroid proximity regardless of local density variation. However, this formulation imposes an inherent convexity constraint. As a result, elongated, curved, nested, and manifold-structured clusters are often represented inadequately, since such geometries cannot generally be captured through linear partitioning alone. Density-based clustering methods address this limitation by enabling the recovery of clusters with arbitrary shape, though they introduce challenges of their own. SPORE is designed to address the geometric rigidity of linear-separation methods while also mitigating two specific limitations commonly associated with density-based alternatives:

\begin{enumerate}
    \item \textbf{Nonlinear separation (Wall 0).} Density-based methods such as DBSCAN \citep{Ester1996} and HDBSCAN \citep{campello2013density, mcinnes2017hdbscan} are capable of recovering clusters with nonlinear and irregular geometric structure, and SPORE preserves this flexibility.
    
    \item \textbf{Variable density (Wall 1).} A single global density threshold is often insufficient for resolving clusters that exist at substantially different density scales. HDBSCAN partially addresses this issue through hierarchical density estimation; however, sparse clusters may still accumulate insufficient stability during extraction, leading to fragmentation or classification as noise. SPORE addresses this limitation through an initial cluster expansion stage that models each cluster's density characteristics directly rather than requiring them to emerge stably from a density hierarchy.
    
    \item \textbf{Low boundary contrast (Wall 2).} When neighboring clusters exhibit similar local density near their shared boundary, density alone becomes a weak discriminative signal regardless of estimation quality. Reducing connectivity thresholds can be sufficient to prevent over-merging, but it often comes at the expense of increased noise classification. In contrast, linear-separability methods such as K-means maintain clear partitions through centroid proximity. Density-based methods typically forgo this property when relaxing the linear separability assumption. SPORE addresses this limitation through a post-expansion, proximity-driven propagation stage designed to maintain discriminative boundary separation while expanding isolated cluster cores into complete clusters.

\end{enumerate}

\paragraph{SPORE.} This paper introduces SPORE (Skeleton Propagation Over Recalibrating Expansions), a density-based clustering method designed to preserve geometric flexibility (Wall~0) while addressing variable density (Wall~1) and low boundary contrast (Wall~2).

In the first phase, clusters expand breadth-first over a $k$-nearest-neighbor graph, admitting neighboring points through a z-score threshold computed from each cluster's evolving distance statistics. Rather than imposing a single global density scale, as in DBSCAN, or traversing a hierarchy of density scales, as in HDBSCAN, SPORE estimates the characteristic density of each cluster independently and restricts expansion to adjacent regions with sufficiently similar local structure. Under conservative deviation tolerances, this phase functions as a skeleton isolation mechanism that preferentially captures high-confidence cluster interiors while intentionally leaving ambiguous boundary regions unresolved as small fragments.

In the second phase, SPORE resolves these fragments through Small-Cluster Reassignment (SCR), a proximity-driven label propagation procedure in which each fragment point is assigned according to the majority label among its $k$ nearest neighbors from established clusters. Unlike density-based separation, this mechanism relies primarily on local geometric proximity: boundary points that lie marginally closer to one cluster tend to inherit more neighboring representatives from that cluster even when density varies smoothly across the boundary region. Because reassignment is performed locally, separability is required only between adjacent regions rather than across the dataset globally.

\paragraph{Contributions.} The primary contributions of this work are as follows:
\begin{itemize}
\item \textbf{SPORE.} A clustering algorithm that preserves nonlinear shape adaptability while addressing key limitations of existing density-based methods through a two-phase design:

    \begin{itemize}
        \item \textbf{The Expansion phase.} An initial stage that preserves Wall~0 shape flexibility while addressing Wall~1 through breadth-first expansion over a $k$-NN graph using online cluster-specific density statistics.
        
        \item \textbf{Small-Cluster Reassignment (SCR).} A second stage that preserves Wall~0 geometric flexibility while addressing Wall~2 through proximity-driven label propagation based on local $k$-NN majority voting. Because reassignment is performed locally rather than with respect to global cluster centroids, cluster geometry can propagate along nonlinear backbone structure without reintroducing a convexity assumption.
    \end{itemize}
    
\item \textbf{A wall-structured benchmark.} A benchmark suite spanning 28 datasets selected to evaluate clustering performance under the Wall~0, Wall~1, and Wall~2 failure modes. The suite additionally includes linearly separable datasets with high inter-cluster proximity, enabling evaluation of both density-based and centroid-based clustering methods under a unified framework.

\item \textbf{A preprocessing evaluation.} An empirical analysis showing that clipped Min-max scaling consistently improves clustering recovery relative to Standard scaling across all evaluated methods while preserving overall comparative performance trends.

\item \textbf{A Python package.} An open-source Python implementation of SPORE, available at \url{https://pypi.org/project/spore-clustering/}.

\end{itemize}

\section{Related Work}
\label{sec:related}

\subsection{Existing Methods}

\paragraph{K-means.} K-means \citep{MacQueen1967} partitions a dataset into $K$ clusters by alternately assigning points to their nearest centroid and recomputing centroids to minimize within-cluster variance. The resulting partition forms a Voronoi tessellation in which cluster boundaries are determined by hyperplanes equidistant between neighboring centroids and are therefore independent of local density structure. This property makes K-means comparatively robust to Wall~2, since centroid proximity yields a decisive partition even when neighboring clusters exhibit similar local density near their boundary. However, this robustness depends on an implicit linear separability assumption, giving rise to Wall~0 limitations. Elongated, curved, and manifold-supported structures are often represented inadequately because Voronoi partitions cannot generally conform to nonlinear cluster geometry. K-means and density-based methods therefore occupy complementary positions: K-means remains effective in settings where density contrast is weak, while density-based methods are more effective when cluster geometry departs substantially from convex structure. This complementarity motivates the two-phase design of SPORE, which separates boundary delineation from density-based cluster discovery.

\paragraph{DBSCAN.} DBSCAN \citep{Ester1996} defines clusters as maximal sets of density-connected points, where density is determined through a fixed neighborhood radius $\varepsilon$ and a minimum neighborhood cardinality parameter, \texttt{MinPts}. The method naturally recovers clusters with arbitrary geometric structure (Wall~0) and identifies sparse observations as noise without requiring a predefined number of clusters. Its principal limitation is Wall~1: a single global $\varepsilon$-\texttt{MinPts} configuration is often insufficient when clusters exist at substantially different density scales. Exposure to Wall~2 follows directly from the connectivity formulation itself. When neighboring clusters exhibit similar local density near their boundary, density connectivity provides little information for determining where one cluster ends and the other begins, often resulting in over-merged partitions. Reducing $\varepsilon$ can partially mitigate this issue by isolating only the densest cluster cores; however, this frequently causes large portions of boundary and peripheral structure to be classified as noise. As a result, preventing over-merging often comes at the expense of discarding substantial fractions of otherwise meaningful data.

\paragraph{HDBSCAN.} HDBSCAN \citep{campello2013density, mcinnes2017hdbscan} extends DBSCAN by incorporating hierarchical density estimation and stability-based cluster extraction. Like DBSCAN, it preserves Wall~0 shape flexibility while improving robustness to Wall~1 by allowing clusters at multiple density scales to emerge within a shared hierarchy. However, the method continues to rely on density connectivity as both its cluster discovery and boundary delineation mechanism. Consequently, Wall~1 is mitigated but not fully resolved: sparse clusters may accumulate insufficient stability as density thresholds increase, leading to fragmentation or absorption prior to extraction. Wall~2 also remains unresolved. When neighboring clusters exhibit similar local density near their boundary, the hierarchy itself contains limited information about the appropriate partition. Conservative extraction settings can reduce over-merging by favoring only the most stable cluster interiors, but this often comes at the expense of fragmenting peripheral structure or classifying substantial portions of the data as noise.

\paragraph{OPTICS.} \citet{Ankerst1999} addressed DBSCAN's Wall~1 
limitation while keeping its shape adaptivity (Wall~0) by replacing the 
fixed-radius partition with a reachability ordering: points are processed in 
order of increasing core distance, and a reachability distance is recorded 
for each point, encoding the density scale at which it connects to the 
growing cluster. The resulting reachability plot represents the full density 
hierarchy of the data, with clusters at different scales appearing as 
distinct valleys rather than being collapsed under a single global 
$\varepsilon$. However, a flat clustering must still be extracted from this 
representation as a post-hoc step, either via a fixed $\varepsilon$ cutoff 
or a heuristic like $\xi$-extraction, which reintroduces a tuning burden and 
leaves Wall~1 only partially resolved. HDBSCAN can be understood as OPTICS 
with a principled answer to the extraction problem: it constructs the same 
density hierarchy via mutual reachability distances and selects clusters by 
optimizing a stability criterion over the condensed tree, removing extraction 
as a separate tuning stage. HDBSCAN therefore serves as the hierarchical 
density baseline in the evaluation in place of OPTICS.

\paragraph{Density Peaks Clustering.} \citet{rodriguez2014} proposed an approach in which cluster centers are identified as local density maxima that are also far from any point of higher density, relying primarily on relative rather than absolute density estimates. This partially addresses Wall~1 by making center identification less sensitive to global density scale, while the density-gradient assignment process preserves non-convex cluster structure (Wall~0). When adjacent clusters share similar density near their boundary, the inward gradient of each cluster's local density field can provide partial robustness to Wall~2 by encouraging boundary points to propagate toward nearby density maxima. However, the method remains sensitive to structural variation in the density landscape. DPC implicitly favors cluster structures organized around dominant density peaks, which can become problematic when clusters are elongated or internally uniform in density. In such settings, minor density fluctuations within a single cluster may generate multiple local maxima, fragmenting one semantic cluster into several regions. Conversely, when neighboring clusters differ substantially in overall density, the assignment cascade may allow a denser cluster to absorb lower-density regions from an adjacent cluster, since each point inherits the label of its nearest neighbor with higher density. Variants of DPC have been proposed to improve center detection under heterogeneous density conditions, but the assignment process itself remains fundamentally density-ordered, leaving these behaviors only partially mitigated.

\paragraph{Shared nearest-neighbor methods.} \citet{jarvis1973} introduced a nonparametric clustering technique based on shared nearest-neighbor similarity, connecting two points when they mutually appear in each other's $k$-NN lists and share at least $k_t$ neighbors. \citet{ertoz2003} extended this into SNN-DBSCAN, redefining pairwise similarity as the count of shared nearest neighbors and then identifying core points and building clusters around them. Shared-neighbor similarity normalizes local connectivity, recovering non-convex shapes (Wall~0) and partially alleviating Wall~1 by allowing clusters at different density scales to form around locally consistent neighborhoods. However, Wall~2 exposure is not eliminated: when adjacent clusters exhibit similar local density near their boundary, shared-neighbor counts may draw support from both clusters, weakening boundary delineation where density contrast is limited.

\paragraph{$k$-NN classifiers and label propagation.} SPORE's Small-Cluster Reassignment (SCR) phase is related to two established frameworks. First, it relates to $k$-nearest neighbor classification \citep{Cover1967}: SCR applies a $k$-NN voting procedure to points contained in small unresolved clusters, assigning each point to the plurality label among its $k$ nearest neighbors drawn from established clusters. Because assignments recursively update the neighborhood label distribution, the procedure also functions as a label propagation process over the $k$-NN graph. This connects naturally to the framework of \citet{zhu2003semi}, who formulated semi-supervised learning as the propagation of labels across a weighted graph according to graph structure. Unlike density-connectivity mechanisms, neighborhood voting does not require an explicit density contrast gradient to produce a decisive assignment. When local neighborhood composition reflects underlying geometric proximity, boundary points can still be assigned consistently even when density varies smoothly across the boundary region (Wall~2). However, unnormalized $k$-NN voting remains sensitive to density imbalance: denser clusters contribute more representatives to a fixed-size neighborhood than geometrically comparable sparse clusters. SPORE addresses this issue by incorporating a density-normalization step into SCR, described in \S~\ref{sec:proposed_method}.

\subsection{Positioning SPORE}
SPORE's two phases can each be situated within the Wall~0--Wall~2 framework introduced in \S~\ref{sec:intro}.

\paragraph{Expansion.} SPORE preserves Wall~0 shape flexibility through graph-based cluster expansion in the density-based paradigm, similarly to DBSCAN and HDBSCAN. Its improvement on Wall~1 arises from replacing a global density criterion with evolving per-cluster statistics. Like DPC, SPORE preferentially seeds clusters in denser regions before expanding them through breadth-first traversal of a $k$-NN graph. However, rather than relying on a shared global density threshold, SPORE evaluates candidate points using a cluster-specific density variance criterion defined by the maximum permitted z-score deviation from the cluster's evolving mean neighbor distance. As a result, each cluster expands according to its own observed density regime rather than a single global density scale.

\paragraph{SCR.} SPORE's SCR phase addresses Wall~2 through proximity-driven $k$-NN majority voting. Points belonging to small unresolved clusters are reassigned to whichever established cluster holds the plurality among their $k$ nearest neighbors, propagating cluster labels locally across the $k$-NN graph. Because voting is determined primarily by geometric proximity rather than density contrast, SCR can preserve discriminative boundaries even in regions where neighboring clusters exhibit similar local density. Unlike centroid-based reassignment, this propagation occurs locally along the recovered cluster backbone structure, allowing nonlinear geometry identified during Expansion to persist throughout reassignment rather than reintroducing a global convexity assumption (Wall~0). Unnormalized $k$-NN voting remains sensitive to density imbalance, however: denser clusters contribute more representatives to a fixed-size neighborhood than geometrically comparable sparse clusters, potentially biasing vote outcomes toward higher-density regions. SPORE addresses this issue by maintaining the z-score filter introduced during Expansion, using it to exclude candidate neighbors for which the reassigned point represents a strong local outlier. As a result, the voting process is restricted primarily to neighboring clusters with comparable local density scale.

\section{Proposed Method}
\label{sec:proposed_method}

\newcommand{\norm}[1]{\left\lVert #1 \right\rVert}

Let $X = \{x_i \in \mathbb{R}^d\}_{i=1}^n$ denote a dataset, and let $\norm{b - a}$ denote the Euclidean distance between points $a, b \in X$.

\subsection{Phase 1: Expansion}

\subsubsection{Density-Ordered Seeding and Temporal Shielding}

Clusters are seeded in ascending order of $k$th-nearest-neighbor distance, placing initial seeds in the densest regions first. Each cluster then expands through breadth-first traversal of the $k$-NN graph. Points are not revisited during expansion, so traversal halts once every neighbor in the active frontier has already been assigned or evaluated. The cluster is then finalized, after which the next densest unassigned point is selected as a new seed. Graph-based expansion in this manner enables SPORE to recover arbitrary geometric structure represented by the $k$-NN graph (Wall~0). Density-ordered seeding combined with non-revisitation allows dense regions to claim nearby points earlier in the expansion process, temporally shielding them from subsequent expansion by sparser clusters and thereby contributing to robustness against Wall~1.

\subsubsection{Density-Variance Filtering}
Not all neighboring points encountered during breadth-first expansion are accepted into the cluster. SPORE filters candidate points according to a density-variance criterion designed to maintain an approximately consistent local density scale within each cluster. This mechanism forms the primary basis for SPORE's Wall~1 robustness.

\paragraph{Evolving Distance Statistics.}
Density variance is estimated through cluster-specific statistics. For each growing cluster $C \subset X$, SPORE maintains the running mean $\mu_C$ and standard deviation $\sigma_C$ of all retained $k$-NN distances accumulated since cluster initialization:
\[
\mu_C = \frac{1}{n_s}\sum_{i=1}^s \sum_{j=1}^{k_i} \delta_{ij},
\qquad
\sigma_C = \sqrt{\frac{1}{n_s}\sum_{i=1}^s \sum_{j=1}^{k_i} \left(\delta_{ij} - \mu_C\right)^2},
\]
where $\delta_{ij}$ denotes the distance to the $j$th nearest neighbor of point $i$, $s$ is the current cluster size, $k_i$ is the number of neighbors accepted when point $i$ was expanded, and $n_s$ is the total number of retained neighbor-distance samples accumulated so far. 

\paragraph{The Density Variance (DV) Filter.}
A candidate neighbor $b$ of the current frontier point $a \in C$ is admitted only if its distance from $a$ remains sufficiently consistent with the cluster's evolving density statistics:
\[
\frac{\norm{b - a} - \mu_C}{\sigma_C} > z
\quad\Longrightarrow\quad
b \notin C.
\]
The parameter $z$ defines the maximum permitted standardized deviation. Unlike DBSCAN's fixed radius $\varepsilon$, the effective inclusion radius adapts continuously as $(\mu_C, \sigma_C)$ evolve, widening in sparse regions and tightening in dense regions according to the cluster's observed local structure.

\paragraph{Online Statistic Updates.}
In practice, density statistics are maintained efficiently through batch-style running averages. Since the variance $\sigma_C^2$ is itself an average of squared deviations, it admits the same weighted update structure as the mean. When $k_{\mathrm{acc}}$ neighbors of a frontier point $i$ are accepted, the statistics are updated according to
\[
\sigma_C^2 \leftarrow \frac{\sigma_C^2 \cdot n_s + \sum_{j=1}^{k_{\mathrm{acc}}}(\delta_{ij} - \mu_C)^2}{n_s + k_{\mathrm{acc}}},
\qquad
\mu_C \leftarrow \frac{\mu_C \cdot n_s + \sum_{j=1}^{k_{\mathrm{acc}}} \delta_{ij}}{n_s + k_{\mathrm{acc}}},
\qquad
n_s \leftarrow n_s + k_{\mathrm{acc}}.
\]

This formulation intentionally forgoes the classical Welford recurrence in favor of treating variance as a direct running average, symmetric in form with the mean update. Empirically, this approximation produces stable expansion behavior. In dense regions, the resulting variance is modestly inflated, slightly relaxing the DV filter and thereby accommodating the short nearest-neighbor distances that might otherwise stall expansion and impair skeleton formation. Initialization of $(\mu_C, \sigma_C)$ to the global statistics $(\mu_\delta, \sigma_\delta)$ with $n_s = 1$ further stabilizes early expansion by providing a well-scaled prior that is progressively overwritten as cluster-specific evidence accumulates.

\subsubsection{The Retention Rate}
Analogous to \texttt{MinPts} in DBSCAN, the parameter $r \in [0,1]$ introduces a local support requirement. After DV filtering, the surviving neighbor set $\Phi_a$ must satisfy
\[
|\Phi_a| \geq \lfloor rk \rfloor,
\]
otherwise expansion from $a$ terminates. This criterion discourages traversal through thin or weakly connected bridge regions between neighboring clusters, even if density is consistent across them.

\subsubsection{Skeleton Formation}
When clusters are adjacent, nested, or separated by weak density contrast at their boundaries, conservative expansion settings allow SPORE to isolate high-confidence interior regions before entering ambiguous boundary structure. As a cluster expands outward from dense seed regions toward sparser peripheral regions, smaller values of $z$ increase the likelihood that candidate points are rejected by the DV filter, while larger retention requirements increase the likelihood that expansion halts at boundary points. The resulting structure consists of compact, high-purity cluster skeletons surrounded by a fragmented set of unresolved points. These skeletons are subsequently propagated across the remaining point cloud during the SCR phase.

\begin{figure}[t]
    \centering
    \begin{minipage}{0.31\textwidth}
        \centering
        \includegraphics[width=\linewidth]{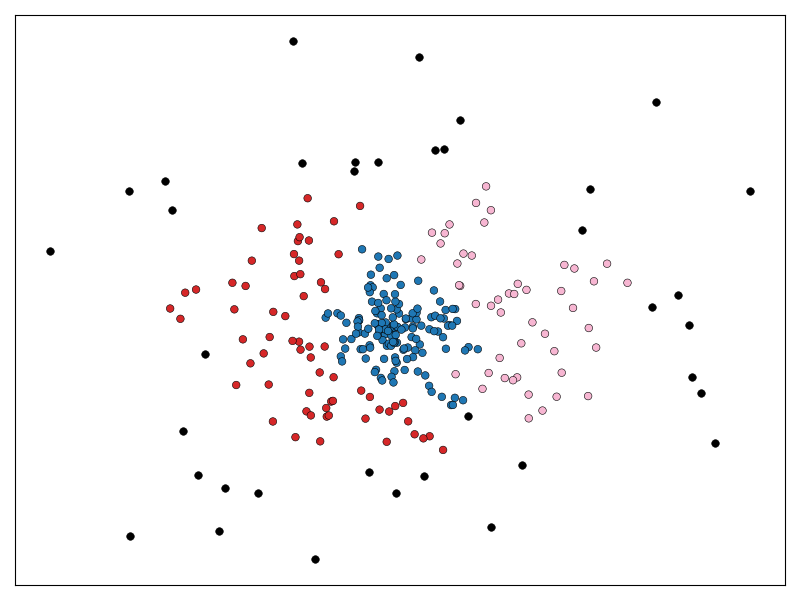}
        \caption*{(a) $z = 1$. The central mass is split into three clusters due to conservative expansion.}
    \end{minipage}\hfill
    \begin{minipage}{0.31\textwidth}
        \centering
        \includegraphics[width=\linewidth]{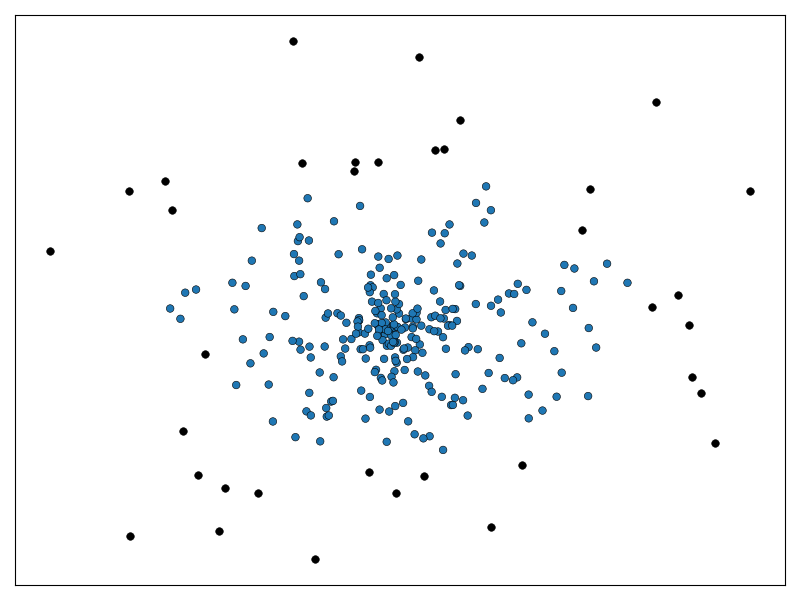}
        \caption*{(b) $z = 2$. The inclusion radius widens, creating one large central cluster.}
    \end{minipage}\hfill
    \begin{minipage}{0.31\textwidth}
        \centering
        \includegraphics[width=\linewidth]{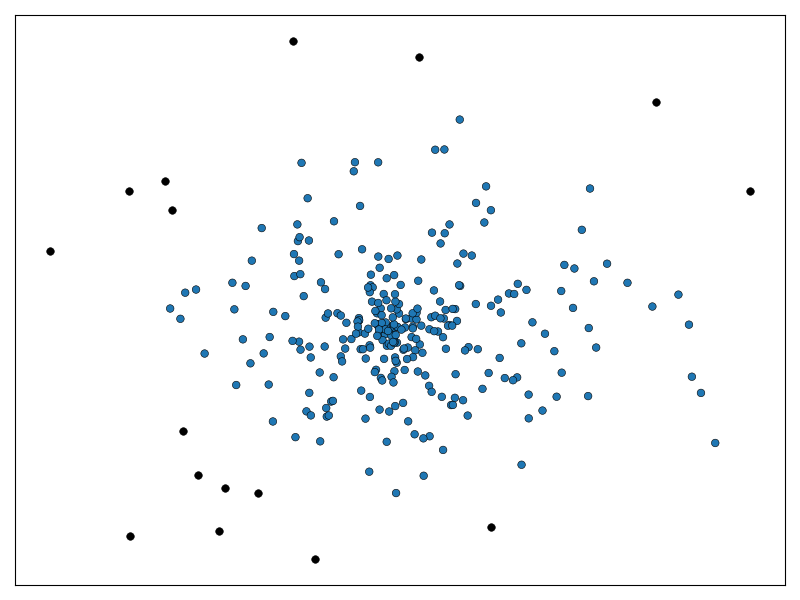}
        \caption*{(c) $z = 3$. The central cluster grows to capture all but the  outermost points.}
    \end{minipage}
    \caption{Effect of the expansion threshold $z$ on a single Gaussian 
    cloud. Small clusters are discarded as noise to isolate the reach of 
    the expansion process. As $z$ increases, the effective inclusion radius 
    $\mu_C + z\,\sigma_C$ grows, admitting progressively sparser regions. 
    At low $z$, the central cluster is small, allowing the sparse regions to its side to become their own clusters; at high $z$, expansion claims 
    the majority of the point cloud. This controls a skeleton--shell tradeoff: conservative 
    $z$ produces tight skeletons that SCR must propagate outward; 
    permissive $z$ produces full clusters directly.}
    \label{fig:gaussian_expansion}
\end{figure}

\FloatBarrier
\begin{algorithm}[t]
\caption{\textsc{SPORE} --- Phase 1: Expansion}
\label{alg:spore_expansion}
\begin{algorithmic}[1]
\STATE \textbf{Input:} Dataset $X=\{x_i \in \mathbb{R}^d\}_{i=1}^n$, expansion threshold $z$, retention rate $r$
\STATE \textbf{Output:} Labels $L$
\STATE Build spatial index over $X$
\STATE Compute $k = O(\log n)$ nearest-neighbor distances $\delta$ for all points
\STATE Initialize priority queue $U$ ordered by increasing $\delta_{ik}$
\STATE Mark all points \texttt{UNVISITED}; set cluster label $c \gets 0$
\WHILE{$U$ not empty}
    \STATE Initialize queue $Q$ with the next \texttt{UNVISITED} point from $U$
    \STATE Initialize $(\mu_C, \sigma_C) \gets (\mu_\delta, \sigma_\delta)$
    \WHILE{$Q$ not empty}
        \STATE Dequeue point $i$; mark $i$ as \texttt{VISITED}; $L_i \gets c$
        \STATE Collect $k_u \leq k$ unvisited neighbors $\Phi_i$ with distances $\delta_i$
        \STATE Remove neighbors with $(\delta_{ij} - \mu_C)/\sigma_C > z$
        \IF{$|\Phi_i| \geq \lfloor rk \rfloor$}
            \STATE Update $(\mu_C, \sigma_C)$ using retained distances
            \STATE Enqueue retained neighbors into $Q$; remove them from $U$
        \ENDIF
    \ENDWHILE
    \STATE $c \gets c + 1$
\ENDWHILE
\STATE \textbf{return} $L$
\end{algorithmic}
\end{algorithm}

\FloatBarrier
\subsection{Phase 2: Small-Cluster Reassignment (SCR)}
\label{sec:scr_phase}
After Expansion, any cluster whose size falls below threshold $M$ is designated a fragment. SCR iteratively resolves these fragments over one or more propagation rounds. During each round, all points currently belonging to small clusters are considered for reassignment, but may only be assigned to established skeleton clusters that are sufficiently represented within their local $k$-NN neighborhood under majority voting. Points surrounded primarily by other fragment points may initially lack a valid reassignment candidate within their neighborhood. However, as neighboring fragment points are progressively absorbed into skeleton clusters, those clusters become reachable to increasingly isolated points through subsequent rounds. As in Expansion, reassignment proceeds in density order so that skeleton propagation follows an ordering analogous to the original cluster expansion process.

\subsubsection{Degeneracy Clamp}
Before SCR begins, $M$ is clamped to $\min(M, |C_2|)$, where
\[
C_2 = \arg\max_{C \in \mathcal{C} \setminus \{C_1\}} |C|
\]
is the second-largest cluster produced during Expansion, and
\[
C_1 = \arg\max_{C \in \mathcal{C}} |C|
\]
is the largest. Without this constraint, excessively large values of $M$ could designate every cluster except $C_1$ as a fragment, causing SCR to absorb the dataset into a single dominant cluster independent of the underlying structure. The clamp ensures that at least two clusters always remain eligible to receive fragment assignments, preserving the minimum nontrivial partition. When $M < |C_2|$, which is the typical operating regime, the clamp has no effect.

\subsubsection{Assignment Procedure}
Let $x$ be a point to assign, $N_k(x)$ its $k$-nearest neighbors, and $L_y$
the label of neighbor $y$. Let $|C_{L_y}|$ denote the post-Expansion size of
the cluster to which $y$ belongs, and let $M$ denote the minimum cluster-size
threshold.

\paragraph{Neighbor filter.}
Three conditions are applied sequentially to obtain the filtered neighbor set
$\widetilde{N}_k(x)$:
\begin{enumerate}
    \item \textbf{Label filter.} Exclude neighbors from $x$'s current cluster:
    $L_y \neq L_x$.
    \item \textbf{Size filter.} Exclude neighbors belonging to clusters whose
    size is below $M$: $|C_{L_y}| \geq M$.
    \item \textbf{Density-normalization gate.} Exclude neighbors whose
    distance from $x$ exceeds $z_\textsc{scr}$ standard deviations above their
    cluster's mean $k$-NN distance:
    \[
        \frac{\|y - x\| - \mu_{C_{L_y}}}
             {\sigma_{C_{L_y}}}
        \leq z_\textsc{scr}.
    \]
\end{enumerate}
Formally:
\[
\widetilde{N}_k(x)
=
\bigl\{
    y \in N_k(x)
    \;:\;
    L_y \neq L_x
    \;\wedge\;
    |C_{L_y}| \geq M
    \;\wedge\;
    \frac{\|y - x\| - \mu_{C_{L_y}}}
         {\sigma_{C_{L_y}}}
    \leq z_\textsc{scr}
\bigr\}.
\]
If $\widetilde{N}_k(x) = \emptyset$, the assignment of $x$ remains unchanged
for that round. A later round may resolve the point if reassignments elsewhere
extend skeleton connectivity into $x$'s local neighborhood. Points that remain
unresolved after all rounds are assigned to noise in the post-SCR step
(\S\ref{sec:pscr}).

\paragraph{Scoring rule.}
For each unique label $\ell$ appearing in
$\widetilde{N}_k(x)$, let
\[
\widetilde{N}_k^{(\ell)}(x)
=
\{\, y \in \widetilde{N}_k(x) : L_y = \ell \,\}.
\]
The candidate cluster score is then defined as the neighbor count:
\[
S_{\ell} = \bigl|\widetilde{N}_k^{(\ell)}(x)\bigr|,
\]
with $S_{\ell} = 0$ when $\widetilde{N}_k^{(\ell)}(x) = \emptyset$.
The final assignment is:
\[
    L_x = \arg\max_{\ell} \, S_{\ell}.
\]

\begin{figure}[t]
    \centering
    \begin{minipage}{0.31\textwidth}
        \centering
        \includegraphics[width=\linewidth]{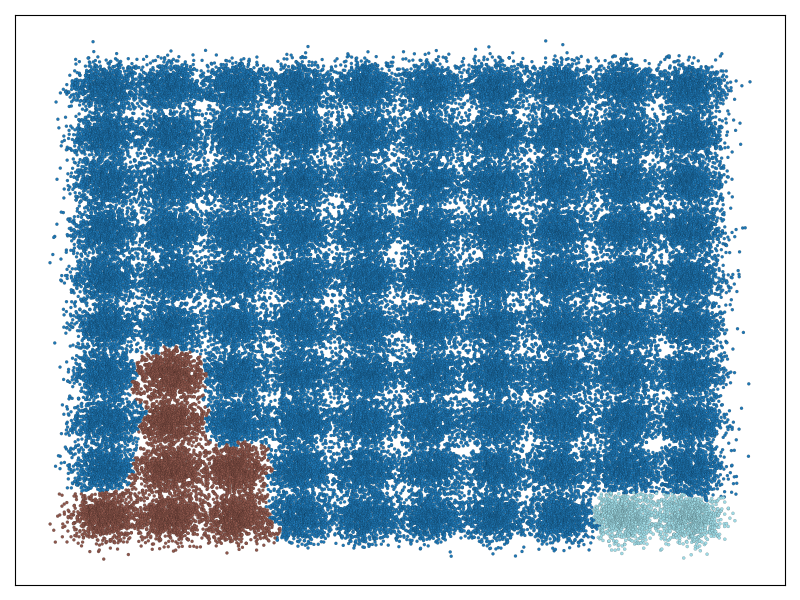}
        \caption*{(a) Default parameters. Clusters merge across low-contrast 
        boundaries, causing most ground-truth clusters to collapse into a 
        single connected component.}
    \end{minipage}\hfill
    \begin{minipage}{0.31\textwidth}
        \centering
        \includegraphics[width=\linewidth]{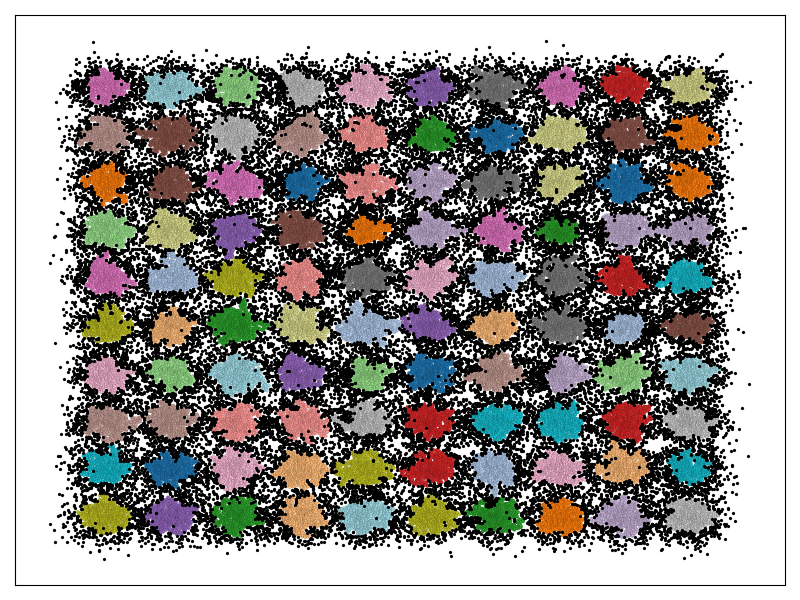}
        \caption*{(b) $r{=}0.125$, \texttt{max\_rounds} = 0. True-cluster 
        skeletons are isolated, but surrounding shell regions are assigned to 
        noise, substantially reducing recovery 
        ($\mathrm{ARI} \approx 0.09$).}
    \end{minipage}\hfill
    \begin{minipage}{0.31\textwidth}
        \centering
        \includegraphics[width=\linewidth]{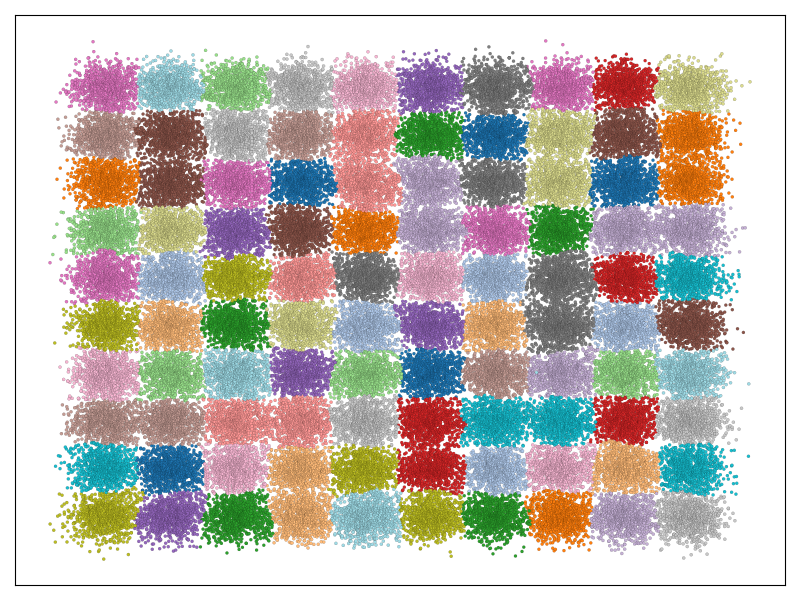}
        \caption*{(c) $r{=}0.125$, \texttt{max\_rounds} = 1. Fragments are 
        resolved through SCR. Shell points are reassigned to their correct 
        clusters despite weak boundary contrast, restoring recovery to 
        $\mathrm{ARI} \approx 0.95$.}
    \end{minipage}
    \caption{SPORE on \texttt{birch1}~\citep{gagolewski2022benchmark}. Panel 
    (a) illustrates the Wall~2 failure mode under default expansion, where 
    excessive propagation across density-connected boundaries produces merged 
    clusters. Panel (b) isolates the skeleton formation mechanism. Increasing 
    the retention rate preserves compact cluster cores, but assigning 
    surrounding shell points to noise substantially degrades recovery. Panel 
    (c) demonstrates the role of SCR: cluster skeletons propagate over nearby 
    fragments through $k$-NN majority voting, producing sharp partitions across 
    low-contrast regions and restoring near-complete recovery.}
    \label{fig:birch1_progression}
\end{figure}

\subsubsection{Post-SCR Noise Assignment}
\label{sec:pscr}
After all propagation rounds complete, any point whose filtered neighbor set
$\widetilde{N}_k(x)$ remained empty throughout the procedure is assigned an
unclustered (noise) label. Persistent exclusion by the density-normalization gate indicates that the point lies outside the density regime of every sufficiently large cluster and is therefore treated as noise.

\FloatBarrier
\begin{algorithm}[t]
\caption{\textsc{SPORE} --- Phase 2: Small-Cluster Reassignment (SCR)}
\label{alg:spore_scr}
\begin{algorithmic}[1]
\STATE \textbf{Input:} Labels $L$ from Phase 1, dataset $X$, min cluster size $M$, $\texttt{max\_rounds} = 10$
\STATE \textbf{Output:} Updated labels $L$

\STATE Apply degeneracy clamp to $M$

\FOR{$\texttt{round} = 1$ \textbf{to} $\texttt{max\_rounds}$}
  \STATE $R \gets \{\,i : |C_{L_i}| < M\,\}$
  \IF{$R = \varnothing$}
    \STATE \textbf{break}
  \ENDIF
  \STATE Sort $R$ by increasing $\delta_{ik}$
  \STATE $\texttt{reassigned} \gets 0$
  \FOR{\textbf{each} $i \in R$}
    \STATE $L_i^{\text{prev}} \gets L_i$
    \STATE Reassign $i$ per scoring rule (\S\ref{sec:scr_phase})
    \IF{$L_i \neq L_i^{\text{prev}}$}
      \STATE $\texttt{reassigned} \gets \texttt{reassigned} + 1$
    \ENDIF
  \ENDFOR
  \IF{$\texttt{reassigned} = 0$}
    \STATE \textbf{break}
  \ENDIF
\ENDFOR

\STATE \textbf{// Post-SCR: assign remaining points to noise}
\FOR{\textbf{each} $i$ with $|C_{L_i}| < M$}
  \STATE $L_i \gets \texttt{NOISE}$
\ENDFOR
\STATE \textbf{return} $L$
\end{algorithmic}
\end{algorithm}

\FloatBarrier
\subsection{Hyperparameter Space Enrichment}
\label{sec:param_enrichment}
SPORE uses several parameters whose raw forms are poorly suited for
dataset-agnostic search. In particular, the expansion threshold $z$, the SCR
gate threshold $z_{\mathrm{scr}}$, and the minimum cluster size $M$ depend
directly on dataset-specific distance distributions or dataset size. These are replaced with bounded, scale-invariant parameterizations that admit fixed search ranges across datasets while preserving data-adaptive behavior. The retention rate $r \in [0,1]$ already satisfies these properties and therefore requires no transformation.

\paragraph{Expansion: Neighborhood Percentile $q_z$.}
$z$ is replaced with a percentile-based parameter $q_z \in [0,100]$ defined over the upper tail of the global $k$th-neighbor distance distribution:
\[
z =
\frac{
Q_{q_z}\!\left(
\{\delta_{ik} : \delta_{ik} \ge \mu_{\delta_{*k}}\}
\right)
-
\mu_{\delta_{*k}}
}{
\sigma_{\delta_{*k}}
},
\]
where $Q_{q_z}$ denotes the $q_z$th percentile operator and
$\delta_{*k}$ denotes the distribution of $k$th-neighbor distances across all
points. This transformation maps the expansion threshold onto the bounded
interval $[0,100]$ independently of dataset geometry. Consequently, a fixed
$q_z$ corresponds to a consistent quantile of the observed distance
distribution rather than an absolute distance scale, improving comparability
across datasets with differing densities and spatial scales.

\paragraph{SCR Gate: Neighborhood Percentile $q_{z_{\mathrm{scr}}}$.}
The SCR gate threshold $z_{\mathrm{scr}}$ is parameterized analogously:
\[
z_{\mathrm{scr}} =
\frac{
Q_{q_{z_{\mathrm{scr}}}}\!\left(
\{\delta_{ik} : \delta_{ik} \ge \mu_{\delta_{*k}}\}
\right)
-
\mu_{\delta_{*k}}
}{
\sigma_{\delta_{*k}}
}.
\]
A default setting $q_{z_{\mathrm{scr}}}=100$ admits all candidate neighbors
except those for which $x$ is an extreme outlier relative to the density regime of the receiving cluster, allowing skeleton propagation to proceed with minimal restriction.

\paragraph{Minimum Cluster Size: Cluster Exponent $M^*$.}
$M$ is replaced with a scale-invariant exponent $M^* \in [0,1]$ such that
\[
M = \lfloor N^{M^*} \rfloor.
\]
This parameterization admits a natural interpretation across its range:
$M^*=0$ gives $M=1$ (SCR inactive), $M^*=0.5$ gives $M=\sqrt{N}$ (a common
heuristic), and $M^*=1$ gives $M=N$ (degenerate).

\section{Time Complexity}
Let $N$ denote the number of data points, $d$ the feature dimensionality,
and $k$ the neighborhood size. Let $C_{\mathrm{nn}}(N,d)$ denote the cost of
a single $k$-NN query: $O(dN)$ under brute-force search and
$O(d\log N)$ when using a spatial index such as a Ball Tree
\citep{omohundro1989balltree}.

\paragraph{Preprocessing.}
SPORE constructs a spatial index over $X$ at cost $O(Nd\log N)$ and computes
the $k$ nearest neighbors of every point at cost
$O(N \cdot C_{\mathrm{nn}}(N,d))$. The resulting neighborhoods and
$k$th-neighbor distances are cached and reused throughout both phases of the
algorithm. Density-ordered seeding requires sorting the $N$ values
$\{\delta_{ik}\}_{i=1}^{N}$, contributing $O(N\log N)$. The percentile-based
parameterizations introduced in \S\ref{sec:param_enrichment} require quantile
computation over the same distance distribution and therefore contribute an
additional $O(N\log N)$ cost. All terms are asymptotically dominated by the
neighbor-search cost.

\paragraph{Expansion phase.}
Each point is expanded at most once. Given a cached neighborhood, Density
Variance filtering, retention-rate evaluation, queue operations, and online
updates of cluster statistics require $O(k)$ work per point. The total cost of
Expansion is therefore
\[
O\!\left(Nk\right).
\]

\paragraph{SCR phase.}
Each propagation round processes only points currently belonging to clusters
smaller than $M$. Reassignment requires filtering and label counting over the
cached neighborhood $\widetilde{N}_k(x)$, which costs $O(k)$ per point.
Because the number of propagation rounds is bounded by the constant parameter
\texttt{max\_rounds}, the total SCR cost is
\[
O\!\left(Nk\right).
\]

\paragraph{Total.}
Combining preprocessing, Expansion, and SCR yields
\[
O\!\left(
  N \cdot C_{\mathrm{nn}}(N,d) + Nk
\right)
=
O\!\left(
  N \cdot \bigl(C_{\mathrm{nn}}(N,d)+k\bigr)
\right).
\]
With an efficient spatial index,
$C_{\mathrm{nn}}(N,d)=O(d\log N)$, giving
\[
O\!\left(N(d\log N+k)\right).
\]
When $k=O(\log N)$, this simplifies to
\[
O(Nd\log N),
\]
with spatial-index construction absorbed into the same asymptotic term.

\FloatBarrier
\section{Representational Capacity Experiment}
\label{sec:rep_cap_experiment}

\subsection{Purpose}

The evaluation serves two objectives. The first is broad: to assess
SPORE's general utility across a diverse range of geometric regimes. The
second is targeted: to empirically stress-test SPORE against the Walls
introduced in \S\ref{sec:intro}. Ground-truth partitions provide a geometry-aware reference against which clustering performance can be evaluated. This enables direct assessment of each representational capacity while distinguishing structural limitations and sensitivities from failures attributable to suboptimal hyperparameter selection.

\subsection{Algorithms}

The following methods are evaluated:

\begin{itemize}
    \item SPORE: Implemented in Python with vectorized NumPy (v2.2.1)
    \item DBSCAN~\citep{Ester1996} (scikit-learn v1.6.1)
    \item HDBSCAN~\citep{campello2013density,mcinnes2017hdbscan} (hdbscan v0.8.40)
    \item DPC~\citep{rodriguez2014}: Implemented in Python with vectorized NumPy (v2.2.1). To automate the clustering process, $k$ is parameterized as the number of cluster centers, which are selected as the top-$k$ points under the $\gamma$-score.
    \item SNN-DBSCAN~\citep{ertoz2003}: Implemented in Python with vectorized NumPy (v2.2.1).
    \item K-means~\citep{MacQueen1967} (scikit-learn v1.6.1)
\end{itemize}

\subsection{Datasets}

Each method is evaluated across 28 datasets spanning synthetic geometric benchmarks, classical tabular datasets, image embeddings, and biological data, covering dimensionalities from 2D to 784D. Geometric characteristics represented include nonconvex structure, manifold support, density variation, nested clusters, and partial overlap. Full descriptions are provided in Appendix~\ref{app:rep_cap_datasets}.

\paragraph{Diagnostic Exemplars.}
The benchmark spans all three wall conditions across low-to-high dimensionality. Wall~0 cases are characterized by nonconvex or manifold-supported cluster structure. Wall~1 cases involve density heterogeneity between neighboring clusters. Wall~2 cases are characterized by low separation combined with low density contrast across cluster boundaries, either through intrinsic geometry, distance concentration in high dimensions, or both. Representative exemplars are summarized in Tables~\ref{tab:exemplars_wall0}--\ref{tab:exemplars_wall2}.

\begin{table}[H]
\centering
\begin{tabular}{lp{8.5cm}}
\toprule
\textbf{Dataset} & \textbf{Characterization} \\
\midrule
\textit{Spiral} (2D) & Three well-separated nonconvex clusters arranged in a spiral configuration, exemplifying shape-based separation. \\
\textit{Smile} (2D) & Multiple well-separated nonconvex components, including ring-like and curved structures arranged in a smile-like pattern. \\
\textit{Aggregation} (2D) & Mostly well-separated globular clusters, some of which are weakly connected by thin bridging regions. \\
\textit{Chainlink} (3D) & Two interlocked toroidal clusters with strong geometric separation. \\
\bottomrule
\end{tabular}
\caption{Wall~0 diagnostic exemplars.}
\label{tab:exemplars_wall0}
\end{table}

\begin{table}[H]
\centering
\begin{tabular}{lp{8.5cm}}
\toprule
\textbf{Dataset} & \textbf{Characterization} \\
\midrule
\textit{Compound} (2D) & Five nonconvex clusters, including a dense cluster embedded within a sparse cluster. \\
\textit{Trapped Lovers} (3D) & Two dense clusters enclosed by a well-separated sparse shell, exhibiting strong density contrast. \\
\textit{Trapped Lovers} (2D) & A PCA-reduced variant that preserves density asymmetry while inducing spatial overlap due to removal of the separating axis. \\
\textit{Isolation} (2D) & Three nested rings with strong density contrast. \\
\textit{Mk4} (3D) & A dense elongated core connected to sparse spiral arms on either end. \\
\bottomrule
\end{tabular}
\caption{Wall~1 diagnostic exemplars.}
\label{tab:exemplars_wall1}
\end{table}

\begin{table}[H]
\centering
\begin{tabular}{lp{8.5cm}}
\toprule
\textbf{Dataset} & \textbf{Characterization} \\
\midrule
\textit{Iris} (4D) & A real-world dataset with three classes: one well-separated and two exhibiting partial overlap. \\
\textit{Mk3} (3D) & Three Gaussian clusters, one well-separated and two overlapping with low density contrast at their boundaries. \\
\textit{G2mg\_128\_30} (128D) & Two Gaussian clusters with intrinsic overlap further compounded by distance concentration. \\
\shortstack{\textit{PBMC} (50D), \textit{Digits} (64D), \\ 
\textit{USPS} (256D), \textit{HAR} (561D)} & Real-world datasets (6--10 classes) where moderate class overlap is amplified by distance concentration in high dimensions. \\
\textit{Fashion} (784D) & Ten classes with high intrinsic inter-class similarity, further exacerbated by high-dimensional distance concentration. \\
\bottomrule
\end{tabular}
\caption{Wall~2 diagnostic exemplars.}
\label{tab:exemplars_wall2}
\end{table}

\FloatBarrier
\subsection{Preprocessing}
\label{sec:preproc_explanation}
For each dataset, features are clipped to $\pm 6\sigma$ and Min-max scaled into $[0,1]$. Standard scaling, though widely used, is not selected because it can distort geometries. For example, a highly discriminating bimodal feature may exhibit high variance, resulting in smaller scale than that of a nondiscriminating unimodal feature. Clustering in this space then over-weights distances along a non-separating axis, making cluster separation more difficult. Clipped Min-max scaling avoids this by reducing the magnitude of strong outliers, preserving within-feature structure, and equalizing feature scale. PCA representation also benefits from this approach: clipping and equalizing feature ranges mitigates scenarios in which principal components become dominated by outlier points, compressing bulk structure in favor of highlighting them.

\subsection{Evaluation Protocol}
For each method and dataset:
\begin{enumerate}
    \item In addition to the default configuration, sample 50 configurations uniformly at random within each method's search bounds (Appendix~\ref{app:search_bounds}). Methods requiring cluster count (K-means, DPC) receive true $k$ in every configuration, so that performance reflects structural advantages and limitations rather than model-selection variance.
    \item Run each configuration under a 120-second time limit, recording ARI, NMI, and wall-clock runtime.
    \item Select the configuration achieving the highest ARI, breaking ties by 
    smallest runtime. For SPORE, this tiebreak carries additional interpretive 
    value: because Phase 2 runtime scales with the number of fragments left 
    unresolved by Phase 1, preferring the fastest equally-accurate configuration helps to assess how permissive expansion can be while retaining maximal accuracy. 
    \item Re-run the selected configuration 10 times to measure solution stability. Due to computational constraints, if a single run of the selected configuration exceeds 2 minutes, no further runs are performed.
\end{enumerate}

All experiments were conducted on a single Intel i7 2.10GHz CPU (12 cores) with 16GB RAM. Multi-core execution was used where supported.

\subsection{Hyperparameter Search Bounds}
Appendix~\ref{app:search_bounds} describes the hyperparameter bounds for each method tested.

\subsection{Metrics}
The primary metric is Adjusted Rand Index (ARI), with $\text{ARI} = 1$ indicating perfect recovery and $\text{ARI} = 0$ indicating chance-level agreement. Normalized Mutual Information (NMI) is reported as a secondary metric. 

\noindent For anytime performance analysis, average highest-achieved ARI is plotted as trial count increases from 1 to 51, characterizing how quickly each method approaches its performance ceiling within the search budget. 

\noindent Runtime is reported as wall-clock runtime for the selected post-tuning configuration, averaged over reruns. 

\noindent Statistical significance is assessed via Wilcoxon signed-rank tests on per-dataset ARI with Holm correction for multiple comparisons. All significance claims in \S\ref{sec:results} refer to Holm-corrected results unless stated otherwise.

\noindent ARI, NMI, and runtime across the full benchmark are reported in Appendix~\ref{app:full_repr_cap_results}.

\subsection{Reproducibility}
SPORE is provided as a clustering package on PyPI [\href{https://pypi.org/project/spore-clustering/}{https://pypi.org/project/spore-clustering/}] and as a repository on GitHub [\href{https://github.com/RandyWAidoo/SPORE}{https://github.com/RandyWAidoo/SPORE}]. The code for the experiment, along with all results and analysis (tables, figures, summaries) is also provided in an associated GitHub repository [\href{https://github.com/RandyWAidoo/SPORE\_Paper}{https://github.com/RandyWAidoo/SPORE\_Paper}].

\FloatBarrier
\section{Experimental Results}
\label{sec:results}

\subsection{Broad Performance}
\subsubsection{Recovery-Runtime Aggregate}
\begin{figure}[ht]
\centering
\includegraphics[width=0.6\textwidth]{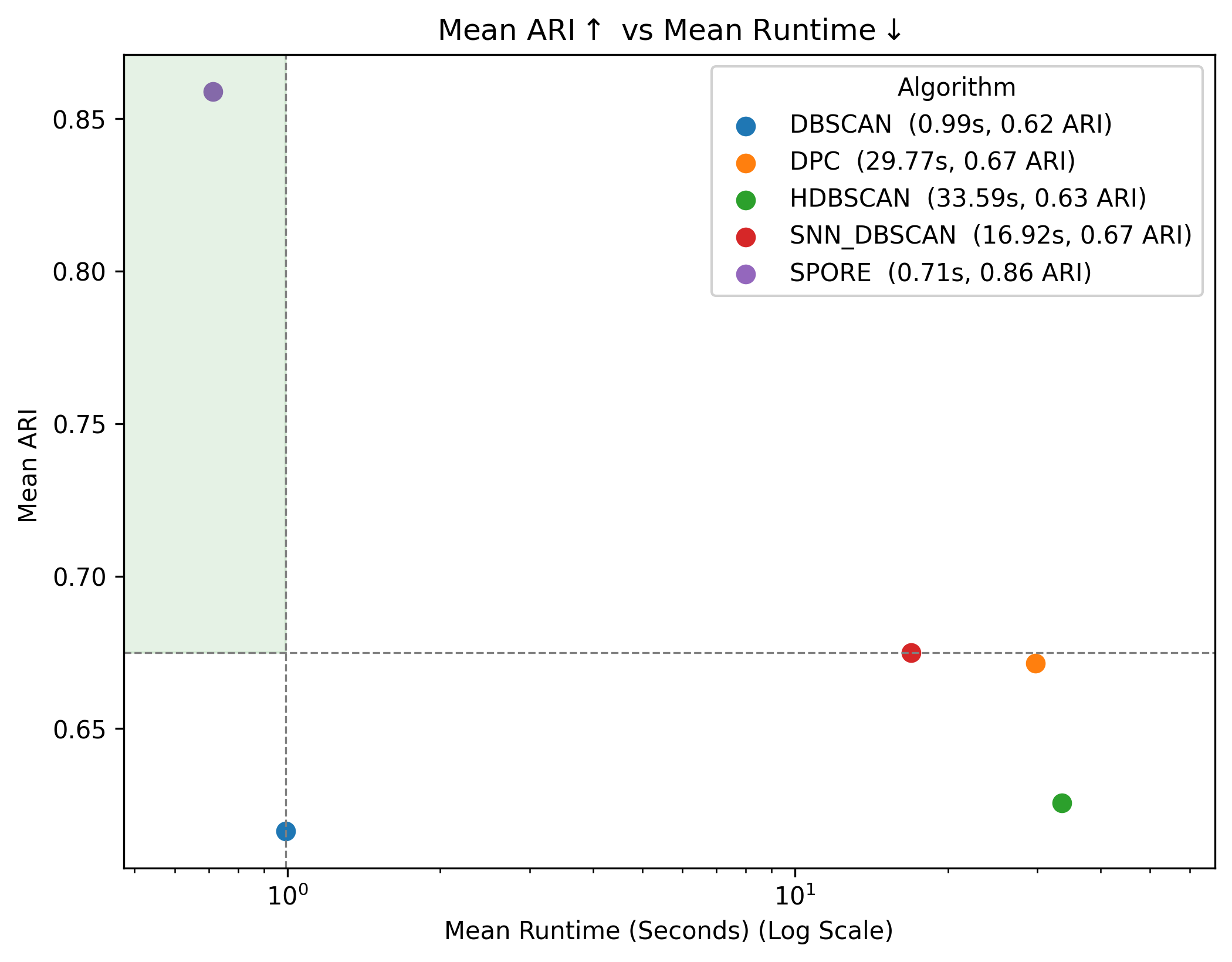}
\caption{Mean ARI versus mean runtime (log scale) across 28 datasets. The shaded region denotes performance exceeding the second-best method on both axes. SPORE is the sole occupant.}
\label{fig:pareto}
\end{figure}

Figure~\ref{fig:pareto} plots mean ARI against mean runtime for the density-based methods. SPORE is the sole Pareto-optimal method, achieving both the highest mean ARI and the lowest mean runtime. SNN-DBSCAN is the closest competitor on ARI but requires roughly 24$\times$ the runtime on average. DBSCAN is the closest baseline in runtime, averaging 0.99\,s compared to SPORE's 0.71\,s, but trails by 0.24 ARI points on average. 

\FloatBarrier
\subsubsection{Recovery}
Averaged across the 28-dataset benchmark, SPORE achieves the highest ARI of any method evaluated. Wilcoxon signed-rank tests with Holm correction confirm that SPORE significantly outperforms every baseline at $p < 0.01$ (Table~\ref{tab:wilcoxon_ari}). The median ARI advantage over DBSCAN, DPC, HDBSCAN, and K-means ranges from 0.10 to 0.22 per dataset. The gap over SNN-DBSCAN is smaller in magnitude (median 0.068) but remains statistically significant, suggesting that shared-neighbor smoothing narrows, but does not eliminate, the advantage associated with the discovery/delineation split.

\begin{table}[ht]
\centering
\caption{Wilcoxon Signed Rank Test for ARI (SPORE vs.\ Others)}
\label{tab:wilcoxon_ari}
\begin{tabular}{lrrrrrc}
\hline
\textbf{Method} & \textbf{N\_pairs} & \textbf{Median Diff} & \textbf{Wilcoxon Stat} & \textbf{$p$-raw} & \textbf{$p$-corrected} & \textbf{Significant} \\
\hline
DBSCAN      & 28 & 0.21605 & 362.0 & 0.000016 & 0.000081 & \checkmark \\
DPC         & 28 & 0.10395 & 269.0 & 0.000337 & 0.000337 & \checkmark \\
HDBSCAN     & 28 & 0.17180 & 241.0 & 0.000101 & 0.000302 & \checkmark \\
KMeans      & 28 & 0.19245 & 326.0 & 0.000066 & 0.000264 & \checkmark \\
SNN\_DBSCAN & 28 & 0.06835 & 299.0 & 0.000120 & 0.000302 & \checkmark \\
\hline
\end{tabular}
\end{table}

\FloatBarrier
\subsubsection{Anytime Analysis}
\label{sec:anytime}

\begin{figure}
    \centering
    \includegraphics[width=0.5\linewidth]{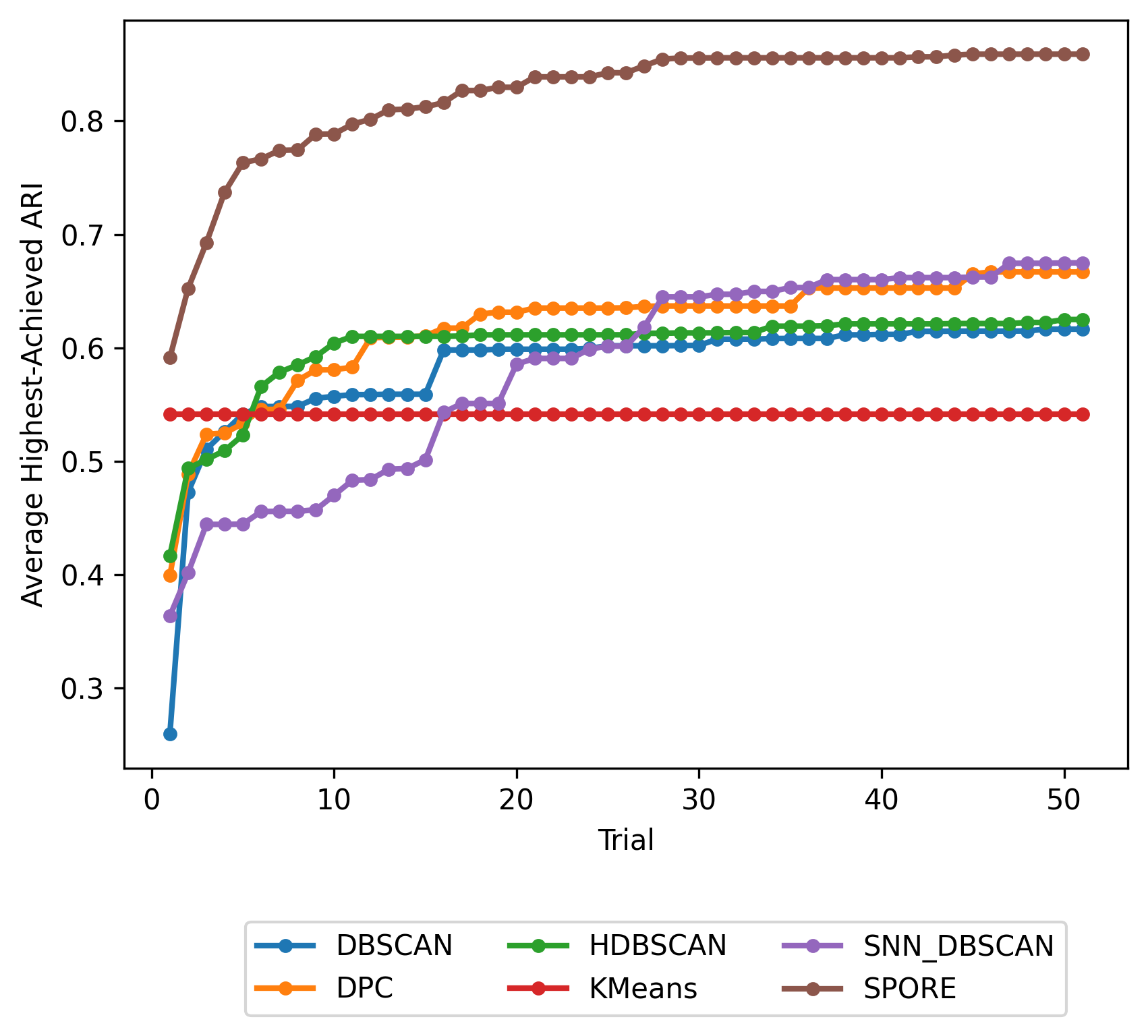}
    \caption{Average highest-achieved ARI (Anytime) performance of each algorithm at each trial count under clipped Min-max scaling. For methods that did not use all 51 trials, the highest score was repeated to 51 trials for symmetry.}
    \label{fig:anytime_curve_z}
\end{figure}

\begin{table}[htbp]
\centering
\caption{Anytime recovery under \textbf{clipped Min-max scaling}. Each cell reports the number of trials required to reach the corresponding average maximum-achieved ARI threshold across 28 datasets. $\infty$ means the threshold was not attained.}
\resizebox{\textwidth}{!}{
\begin{tabular}{l*{15}{c} S S}
\toprule
Algorithm & 0.3 & 0.35 & 0.4 & 0.45 & 0.5 & 0.55 & 0.6 & 0.65 & 0.7 & 0.75 & 0.8 & 0.85 & 0.9 & 0.95 & 1.0 & {Max} & {Default} \\
\midrule
SPORE       & 1 & 1 & 1 & 1 & 1 & 1 & 2 & 2 & 4 & 5 & 12 & 28 & $\infty$ & $\infty$ & $\infty$ & 0.8590 & 0.5912 \\
SNN\_DBSCAN & 1 & 1 & 2 & 6 & 15 & 17 & 25 & 35 & $\infty$ & $\infty$ & $\infty$ & $\infty$ & $\infty$ & $\infty$ & $\infty$ & 0.6749 & 0.3634 \\
DPC         & 1 & 1 & 2 & 2 & 3 & 8 & 12 & 36 & $\infty$ & $\infty$ & $\infty$ & $\infty$ & $\infty$ & $\infty$ & $\infty$ & 0.6669 & 0.3990 \\
HDBSCAN     & 1 & 1 & 1 & 2 & 3 & 6 & 10 & $\infty$ & $\infty$ & $\infty$ & $\infty$ & $\infty$ & $\infty$ & $\infty$ & $\infty$ & 0.6249 & 0.4165 \\
DBSCAN      & 2 & 2 & 2 & 2 & 3 & 9 & 25 & $\infty$ & $\infty$ & $\infty$ & $\infty$ & $\infty$ & $\infty$ & $\infty$ & $\infty$ & 0.6164 & 0.2592 \\
KMeans      & 1 & 1 & 1 & 1 & 1 & $\infty$ & $\infty$ & $\infty$ & $\infty$ & $\infty$ & $\infty$ & $\infty$ & $\infty$ & $\infty$ & $\infty$ & 0.5412 & 0.5412 \\
\bottomrule
\end{tabular}
}
\label{tab:trials_anytime_ari}
\end{table}

\begin{table}[ht]
\centering
\caption{Anytime recovery under \textbf{Standard scaling}. Each cell reports the number of trials required to reach the corresponding average maximum-achieved ARI threshold across 28 datasets. $\infty$ means the threshold was not attained.}
\label{tab:anytime_std}
\resizebox{\textwidth}{!}{%
\begin{tabular}{lrrrrrrrrrrrrrrrrr}
\toprule
Algorithm
  & 0.30 & 0.35 & 0.40 & 0.45 & 0.50 & 0.55 & 0.60 & 0.65 & 0.70 & 0.75 & 0.80 & 0.85 & 0.90 & 0.95 & 1.00
  & Max & Default \\
\midrule
SPORE       & 1 & 1 & 1 & 1 & 1 & 1 & 2 & 3 & 5 & 6 & 13 & $\infty$ & $\infty$ & $\infty$ & $\infty$ & 0.8376 & 0.5940 \\
SNN\_DBSCAN & 1 & 2 & 3 & 6 & 11 & 27 & 31 & 47 & $\infty$ & $\infty$ & $\infty$ & $\infty$ & $\infty$ & $\infty$ & $\infty$ & 0.6514 & 0.3017 \\
DPC         & 1 & 1 & 1 & 2 & 3 & 4 & 13 & $\infty$ & $\infty$ & $\infty$ & $\infty$ & $\infty$ & $\infty$ & $\infty$ & $\infty$ & 0.6302 & 0.4175 \\
HDBSCAN     & 1 & 1 & 1 & 2 & 5 & 7 & 34 & $\infty$ & $\infty$ & $\infty$ & $\infty$ & $\infty$ & $\infty$ & $\infty$ & $\infty$ & 0.6092 & 0.4061 \\
DBSCAN      & 2 & 2 & 2 & 2 & 4 & 16 & 49 & $\infty$ & $\infty$ & $\infty$ & $\infty$ & $\infty$ & $\infty$ & $\infty$ & $\infty$ & 0.6004 & 0.2691 \\
KMeans      & 1 & 1 & 1 & 1 & 1 & $\infty$ & $\infty$ & $\infty$ & $\infty$ & $\infty$ & $\infty$ & $\infty$ & $\infty$ & $\infty$ & $\infty$ & 0.5266 & 0.5266 \\
\bottomrule
\end{tabular}%
}
\end{table}

\paragraph{Anytime performance.}
Table~\ref{tab:trials_anytime_ari} functions as a trial-budget ablation:
each column reports the minimum budget required to reach the corresponding
average best-achieved ARI threshold, characterizing how performance scales
with evaluation cost. At a single-trial budget, SPORE's default ARI of
$\approx 0.59$ approaches the 50-trial maximum of baselines
($\approx 0.62\text{--}0.67$). By 12 trials, SPORE reaches
$\text{ARI} \geq 0.80$, which is a threshold no baseline attains at any
budget. SNN-DBSCAN, the strongest baseline, requires 35 trials to reach
0.65 and is bounded below 0.70. The gap is stable across all budget levels:
SPORE dominates at 1 trial, at 10, and at 50. These results address the
unequal trial counts introduced by runtime constraints on larger datasets:
SPORE surpasses every baseline's 50-trial maximum within five trials, indicating that differences in realized trial counts are unlikely to explain the observed performance gap.

\paragraph{Effect of preprocessing.}
Table~\ref{tab:anytime_std} reports anytime performance under standard
scaling as a preprocessing sensitivity check. All methods decline in maximum
ARI relative to each method's own clipped Min-max ceiling. The declines are
comparable across methods, and all pairwise gaps between SPORE and baselines
are preserved. This supports the rationale for clipped Min-max scaling given in \S\ref{sec:preproc_explanation}. Performance declines under Standard scaling for all methods, consistent with the view that variance normalization can weaken discriminating geometric structure across datasets.

\FloatBarrier
\subsection{Strengths and Weaknesses by Method}

\begin{table}[ht]
\centering
\setlength{\tabcolsep}{2pt}
\begin{tabular}{lcccccc}
\toprule
\textbf{Dataset} & \textbf{DBSCAN} & \textbf{DPC} & \textbf{HDBSCAN} & \textbf{KMeans} & \textbf{SNN-DBSCAN} & \textbf{SPORE} \\
\midrule
\multicolumn{7}{l}{\textit{Wall~0}} \\[2pt]
Spiral              & \textbf{1.0} & \textbf{1.0} & \textbf{1.0} & $-$0.01 & \textbf{1.0} & \textbf{1.0} \\
Smile               & 0.97 & 0.95 & \textbf{1.0} & $0.48 \pm 0.06$ & 0.99 & \textbf{1.0} \\
Chainlink           & \textbf{1.0} & \textbf{1.0} & \textbf{1.0} & 0.09 & \textbf{1.0} & \textbf{1.0} \\
Aggregation         & 0.97 & \textbf{1.0} & 0.91 & $0.73 \pm 0.01$ & 0.90 & 0.99 \\
\midrule
\multicolumn{7}{l}{\textit{Wall~1}} \\[2pt]
Compound            & 0.87 & 0.82 & 0.84 & $0.59 \pm 0.01$ & 0.86 & \textbf{0.98} \\
Trapped Lovers (3D) & \textbf{1.0} & 0.49 & \textbf{1.0} & 0.15 & \textbf{1.0} & \textbf{1.0} \\
Trapped Lovers (2D) & 0.53 & 0.32 & 0.64 & 0.15 & 0.00 & \textbf{0.85} \\
Isolation           & 0.99 & 0.02 & \textbf{1.0} & $-$0.00 & \textbf{1.0} & \textbf{1.0} \\
Mk4                 & 0.61 & 0.76 & \textbf{1.0} & 0.41 & \textbf{1.0} & \textbf{1.0} \\
\midrule
\multicolumn{7}{l}{\textit{Wall~2}} \\[2pt]
Iris                & 0.62 & \textbf{0.89} & 0.57 & 0.72 & 0.88 & \textbf{0.89} \\
Mk3                 & 0.54 & 0.87 & 0.57 & \textbf{0.88} & 0.57 & 0.85 \\
G2mg\_128\_30       & 0.00 & 0.11 & 0.01 & \textbf{0.95} & 0.00 & 0.89 \\
PBMC                & 0.27 & 0.35 & 0.05 & $0.75 \pm 0.11$ & 0.61 & \textbf{0.81} \\
Digits              & 0.41 & 0.54 & 0.65 & $0.66 \pm 0.01$ & 0.78 & \textbf{0.85} \\
USPS                & 0.15 & 0.25 & 0.10 & $0.54 \pm 0.02$ & 0.36 & \textbf{0.64} \\
HAR                 & 0.32 & 0.46 & 0.29 & 0.46 & 0.28 & \textbf{0.60} \\
Fashion             & 0.05 & 0.13 & 0.02 & $\mathbf{0.35} \pm 0.01$ & 0.00 & 0.23 \\
\bottomrule
\end{tabular}
\caption{ARI scores on diagnostic exemplars. Bold indicates the maximum per row. Standard deviations below 0.005 are omitted.}
\label{tab:exemplar_ari}
\end{table}

\subsubsection{DBSCAN}
\paragraph{Wall~0.}
DBSCAN performs strongly across all Wall~0 exemplars, achieving ARI values of 1.0 on \textit{Spiral} and \textit{Chainlink}, 0.97 on \textit{Smile}, and 0.97 on \textit{Aggregation}. This behavior reflects DBSCAN's core representational advantage: clusters are defined through density connectivity rather than centroidal structure, allowing nonconvex and manifold-supported geometries to be recovered naturally. All four exemplars contain well-separated connected density regions, providing exactly the conditions under which DBSCAN is most effective.

\paragraph{Wall~1.}
Performance remains strong when density asymmetry is accompanied by clear geometric separation. DBSCAN achieves near-perfect recovery on \textit{Isolation} (0.99) and perfect recovery on \textit{Trapped Lovers} (3D, 1.0), where dense structures remain spatially distinct despite substantial density differences. Recovery also remains high on \textit{Compound} (0.87), indicating moderate robustness to heterogeneous density. However, performance deteriorates when density asymmetry is coupled with overlap. On \textit{Trapped Lovers} (2D), PCA projection removes the separating axis while preserving the density imbalance, reducing ARI to 0.53. A similar decline occurs on \textit{Mk4} (0.61), where dense and sparse regions are connected within a single elongated structure. These results illustrate a central limitation of DBSCAN: a single global density threshold can accommodate density variation or overlap individually, but becomes increasingly difficult to tune when both occur simultaneously.

\paragraph{Wall~2.}
Wall~2 exposes the principal representational limitation of DBSCAN. Performance declines substantially across all exemplars, falling to 0.62 on \textit{Iris}, 0.54 on \textit{Mk3}, 0.27 on \textit{PBMC}, 0.41 on \textit{Digits}, 0.15 on \textit{USPS}, 0.32 on \textit{HAR}, and 0.05 on \textit{Fashion}. The most severe failure occurs on \textit{G2mg\_128\_30}, where ARI collapses to 0.00. These datasets combine low separation with weak density contrast, often in dimensions where distance concentration further reduces the distinction between intra- and inter-cluster neighborhoods. Under these conditions, DBSCAN's notion of cluster identity becomes increasingly ambiguous: density-connected regions merge or fragment depending on parameter selection. The resulting pattern is not an isolated failure on particular datasets, but a systematic degradation across the full range of Wall~2 exemplars.

\subsubsection{HDBSCAN}
\paragraph{Wall~0.}
HDBSCAN performs strongly across all Wall~0 exemplars, achieving perfect recovery on \textit{Spiral} (1.0), \textit{Smile} (1.0), and \textit{Chainlink} (1.0), while maintaining high performance on \textit{Aggregation} (0.91). As with DBSCAN, this reflects the suitability of density-connected representations for nonconvex and manifold-supported structure. Because Wall~0 datasets contain well-separated density regions, cluster persistence remains stable across a wide range of density thresholds, allowing HDBSCAN's hierarchical selection procedure to recover the correct partition reliably.

\paragraph{Wall~1.}
HDBSCAN handles several forms of density asymmetry more effectively than DBSCAN. It achieves perfect recovery on \textit{Isolation} (1.0), \textit{Mk4} (1.0), and \textit{Trapped Lovers} (3D, 1.0), demonstrating strong robustness to nested structure and large density variation. By modeling cluster structure across multiple density scales, HDBSCAN is substantially less sensitive to heterogeneous density than DBSCAN. However, these advantages are not universal. On \textit{Compound}, HDBSCAN achieves an ARI of 0.84, slightly below DBSCAN's 0.87, indicating that density hierarchy alone does not fully resolve embedded cluster structure. Performance declines further on \textit{Trapped Lovers} (2D) (0.64), where projection-induced overlap removes the separating axis while preserving density asymmetry. Together, these results suggest that hierarchical density estimation mitigates, but does not eliminate, the challenges posed by density asymmetry and overlap.

\paragraph{Wall~2.}
Although HDBSCAN extends DBSCAN through multi-scale density estimation, its performance deteriorates substantially across Wall~2 exemplars. Recovery falls to 0.57 on \textit{Iris}, 0.57 on \textit{Mk3}, 0.05 on \textit{PBMC}, 0.65 on \textit{Digits}, 0.10 on \textit{USPS}, 0.29 on \textit{HAR}, and 0.02 on \textit{Fashion}. On \textit{G2mg\_128\_30}, performance collapses to 0.01. These datasets are characterized by weak density contrast and substantial overlap, often compounded by distance concentration in high dimensions. Under such conditions, the persistence criterion that drives HDBSCAN's cluster selection becomes increasingly uninformative because true clusters do not manifest as strongly differentiated density structures. Consequently, while HDBSCAN substantially expands the range of density heterogeneity it can accommodate relative to DBSCAN, it remains fundamentally constrained by the availability of persistent density-based cluster structure.

\subsubsection{DPC}
\paragraph{Wall~0.}
DPC performs strongly across the Wall~0 exemplars, achieving perfect recovery on \textit{Spiral} (1.0), \textit{Chainlink} (1.0), and \textit{Aggregation} (1.0), while remaining highly accurate on \textit{Smile} (0.95). These results indicate that nonconvexity alone is not problematic for DPC. Once cluster centers are correctly identified, the density-peak assignment procedure is capable of recovering a wide range of geometric structures, including manifolds and nonconvex connected components. The strong Wall~0 performance demonstrates that DPC's representational limitations arise primarily from center identification rather than cluster shape.

\paragraph{Wall~1.}
DPC exhibits highly variable performance across Wall~1 exemplars. It performs reasonably on \textit{Compound} (0.82) and \textit{Mk4} (0.76), where distinct density maxima remain identifiable despite substantial density heterogeneity. However, performance collapses on \textit{Trapped Lovers} (3D) (0.49), \textit{Trapped Lovers} (2D) (0.32), and especially \textit{Isolation} (0.02). Unlike DBSCAN and HDBSCAN, DPC does not rely on density connectivity. Its success depends on the existence of distinct density peaks and reliable density gradients leading toward them. Wall~1 datasets often violate this assumption. In nested structures such as \textit{Isolation}, the densest region dominates the density landscape, causing points from multiple ground-truth clusters to converge toward the same peak. Similarly, in \textit{Trapped Lovers}, the sparse enclosing shell lacks a strong local maximum of its own, making it difficult to recover as a separate cluster. These results indicate that DPC remains vulnerable when density asymmetry obscures or eliminates the distinct peaks required for center identification.

\paragraph{Wall~2.}
DPC remains competitive on several Wall~2 exemplars, achieving 0.89 on \textit{Iris}, 0.87 on \textit{Mk3}, 0.54 on \textit{Digits}, and 0.46 on \textit{HAR}. Unlike density-connectivity methods, DPC assigns points through a sequence of nearest-neighbor decisions toward higher-density points, terminating at density peaks selected via a global argmax criterion. Consequently, DPC can remain effective even when density contrast is weak, provided density gradients still point reliably toward the correct cluster centers. This is most evident on \textit{Mk3}, where overlapping Gaussian clusters retain coherent gradients despite weak separation. Performance deteriorates, however, as distance concentration weakens the gradient signal itself, falling to 0.35 on \textit{PBMC}, 0.25 on \textit{USPS}, 0.13 on \textit{Fashion}, and 0.11 on \textit{G2mg\_128\_30}. DPC therefore occupies an intermediate position between density-connectivity and centroid-based methods: more robust to weak separation than conventional density clustering, but still dependent on the existence of coherent density gradients leading toward identifiable cluster centers.

\subsubsection{SNN-DBSCAN}
\paragraph{Wall~0.}
SNN-DBSCAN performs strongly across all Wall~0 exemplars, achieving perfect recovery on \textit{Spiral} (1.0) and \textit{Chainlink} (1.0), 0.99 on \textit{Smile}, and 0.90 on \textit{Aggregation}. As with DBSCAN, nonconvexity itself presents little difficulty when cluster structure is reflected in local neighborhood connectivity. Replacing Euclidean distance with shared-neighbor similarity preserves the representational advantages of density-based clustering while remaining robust to manifold-supported geometries.

\paragraph{Wall~1.}
SNN-DBSCAN generally performs well under density asymmetry, achieving perfect recovery on \textit{Isolation} (1.0), \textit{Mk4} (1.0), and \textit{Trapped Lovers} (3D) (1.0), while maintaining strong performance on \textit{Compound} (0.86). These results suggest that shared-neighbor similarity reduces sensitivity to absolute density variation by emphasizing neighborhood agreement rather than local point density alone. The principal exception is \textit{Trapped Lovers} (2D), where performance collapses to 0.00 following PCA projection. Once overlap is introduced, the local neighborhoods of the two dense clusters become heavily intermixed, eliminating the shared-neighbor structure required for reliable separation.

\paragraph{Wall~2.}
SNN-DBSCAN outperforms DBSCAN and HDBSCAN across most Wall~2 exemplars, achieving 0.88 on \textit{Iris}, 0.78 on \textit{Digits}, 0.61 on \textit{PBMC}, and 0.36 on \textit{USPS}. By operating on shared-neighbor similarity rather than raw distance, the method benefits from the relative robustness of $k$-NN structure under weak separation and distance concentration. The shared-neighbor criterion also reduces overmerging by requiring strong local neighborhood agreement before regions become connected, making cluster boundaries more resistant to weak inter-cluster proximity. However, this advantage diminishes as overlap increases. Recovery falls to 0.57 on \textit{Mk3}, 0.28 on \textit{HAR}, and 0.00 on both \textit{G2mg\_128\_30} and \textit{Fashion}. In these regimes, points on opposite sides of a cluster boundary increasingly share the same neighbors, causing shared-neighbor affinity itself to become non-discriminative. SNN-DBSCAN therefore extends the useful range of density-based clustering under Wall~2, but remains fundamentally dependent on the existence of separable local neighborhood structure.

\subsubsection{K-means}
\paragraph{Wall~0.}
K-means performs poorly across the Wall~0 exemplars, achieving ARI values of $-0.01$ on \textit{Spiral}, 0.48 on \textit{Smile}, and 0.09 on \textit{Chainlink}. The sole exception is \textit{Aggregation} (0.73), whose predominantly globular structure is more compatible with centroid-based partitioning. These results reflect a fundamental representational limitation: K-means partitions space into Voronoi cells around cluster centroids and therefore cannot naturally represent nonconvex or manifold-supported structure. As a result, performance degrades whenever cluster membership cannot be expressed through proximity to a central prototype.

\paragraph{Wall~1.}
Performance remains weak across the Wall~1 exemplars, with ARI values of 0.59 on \textit{Compound}, 0.15 on both \textit{Trapped Lovers} variants, $-0.00$ on \textit{Isolation}, and 0.41 on \textit{Mk4}. While K-means does not rely on density connectivity, density asymmetry can still influence clustering indirectly by shifting centroid locations toward denser regions. Combined with the nested, shell-like, and highly nonconvex geometries characteristic of Wall~1, this often produces decision boundaries that are poorly aligned with the true partition. Consequently, the centroidal representation underlying K-means remains fundamentally mismatched to the structures these datasets were designed to expose.

\paragraph{Wall~2.}
Wall~2 is the regime in which K-means performs most strongly. The method achieves the highest ARI on \textit{Mk3} (0.88), \textit{G2mg\_128\_30} (0.95), and \textit{Fashion} (0.35), while remaining competitive on \textit{Iris} (0.72), \textit{PBMC} (0.75), \textit{Digits} (0.66), \textit{USPS} (0.54), and \textit{HAR} (0.46). Unlike density-based methods, K-means does not require density contrast, connectivity, or neighborhood agreement. Instead, clustering is driven entirely by repeated nearest-centroid assignments. The iterative averaging and argmin mechanism effectively denoises cluster centroid estimates and allows weak separation signals to be amplified into coherent partitions. Consequently, K-means is largely unaffected by the collapse of density contrast that characterizes Wall~2, provided clusters remain approximately representable by their centroids. The resulting pattern is complementary to that observed for density-based methods: K-means excels precisely where density-based representations begin to lose discriminative power.

\subsubsection{SPORE}
\paragraph{Wall~0.}
SPORE achieves perfect recovery on \textit{Spiral}, \textit{Smile}, and \textit{Chainlink}, and 0.99 on \textit{Aggregation}. These results are expected: density-variance-restricted breadth-first expansion over a $k$-NN graph naturally follows arbitrary connected shapes, including nonconvex and manifold-supported structure. Consequently, SPORE retains the classical representational strengths of density-based clustering, recovering all Wall~0 exemplars with performance comparable to the strongest density-based baselines.

\paragraph{Wall~1.}
SPORE performs strongly across all Wall~1 exemplars, achieving 0.98 on \textit{Compound}, perfect recovery on \textit{Trapped Lovers} (3D), \textit{Isolation}, and \textit{Mk4}, and 0.85 on \textit{Trapped Lovers} (2D). Unlike existing density-based methods, SPORE treats density scale as a cluster-specific property rather than a global constraint. During discovery, expansion is governed by a density variance criterion learned from the evolving cluster itself, allowing dense and sparse structures to be evaluated relative to their own local statistics. This directly addresses the central challenge posed by Wall~1: density asymmetry between neighboring clusters. Whereas other methods approach density asymmetry indirectly through shared-neighbor smoothing, hierarchical density estimation, or globally tuned density thresholds, SPORE learns and applies density tolerances independently for each cluster as it forms. The resulting adaptiveness is particularly evident on \textit{Trapped Lovers} (2D), where projection-induced overlap reduces the effectiveness of conventional density cues. Although recovery is no longer perfect, SPORE substantially exceeds the strongest baseline (0.85 versus 0.64), indicating that cluster-specific density modeling remains informative even when density asymmetry and overlap occur simultaneously.

\paragraph{Wall~2.}
SPORE achieves the highest ARI on \textit{PBMC} (0.81), \textit{Digits} (0.85), \textit{USPS} (0.64), and \textit{HAR} (0.60), while remaining competitive with the strongest methods on \textit{Iris} (0.89), \textit{Mk3} (0.85), \textit{G2mg\_128\_30} (0.89), and \textit{Fashion} (0.23). Unlike conventional density-based methods, SPORE does not require density contrast to remain informative throughout an entire cluster. Instead, discovery only requires that the densest points of a cluster remain concentrated within its interior and sufficiently separated from neighboring clusters to form a reliable skeleton. This requirement is often substantially weaker than maintaining a coherent density gradient or density-connected region across the full cluster. Because $k$-NN structure remains marginally reliable under moderate overlap and distance concentration, particularly within cluster interiors, skeletons can often still be recovered even after conventional density cues begin to degrade. Once established, SCR propagates these interior assignments outward through iterative nearest-neighbor preference. Points near reliable cluster cores are assigned first, while ambiguous points remain unresolved until later rounds, allowing weak or incomplete gradients to be bridged incrementally rather than requiring a coherent density signal across the entire cluster. When multiple cluster cores are visible, assignment is gated by density consistency and governed primarily by local neighborhood preference rather than relative core density, reducing sensitivity to both density asymmetry and low density contrast during delineation. The resulting behavior is most evident on high-dimensional datasets such as \textit{PBMC}, \textit{Digits}, \textit{USPS}, and \textit{HAR}, where SPORE substantially outperforms all baselines. Performance nevertheless declines on \textit{Fashion} and remains below K-means on \textit{Mk3} and \textit{G2mg\_128\_30}, where the convexity assumption is most aligned with the underlying geometry and global centroid assignment remains most resilient.

\paragraph{The Fish Dataset.}
SPORE's performance on \textit{Fish} (Appendix~\ref{app:full_repr_cap_results}) is an irregular instance of under-performance relative to baselines, with ARI dropping to 0.60 while several density-based methods achieve approximately 0.86. This behavior is not a fundamental representational failure, but rather a consequence of a feature-construction artifact: the inclusion of the dependent coordinate $z=y/x$ induces substantial intra-cluster density variation. As a result, individual clusters fragment into multiple dense modes separated by low-density gaps that create weakly connected subregions within the $k$-NN graph. The issue can be resolved either by removing the engineered feature or by adopting a more permissive expansion regime. Increasing \texttt{expansion\_neighbors} to 32 and \texttt{z\_percentile} to 92 increases graph connectivity and allows density-separated modes within the same cluster to be bridged into a common skeleton family, restoring ARI to 0.86. This result highlights a specific sensitivity of SPORE: while robust to density asymmetry and weak density contrast, it remains dependent on sufficient $k$-NN graph connectivity for assignment signals to propagate between regions belonging to the same cluster.

\FloatBarrier
\subsection{The Expansion--SCR Continuum}
\begin{table}[ht]
\centering
\setlength{\tabcolsep}{4pt}
\begin{tabular}{lccccc}
\toprule
\textbf{Dataset} & \textbf{N} & \textbf{D'} & $\boldsymbol{q_z}$ & $\boldsymbol{r}$ & $\boldsymbol{q_z/100 - r}$ \\
\midrule
\multicolumn{6}{l}{\textit{Wall~0}} \\[2pt]
Spiral               & 312   &   2 & 90.88 & 0.18 &  0.73 \\
Smile                & 1000  &   2 & 62.17 & 0.18 &  0.44 \\
Aggregation          & 788   &   2 & 78.61 & 0.36 &  0.43 \\
Chainlink            & 1000  &   3 & 91.01 & 0.08 &  0.83 \\
\midrule
\multicolumn{6}{l}{\textit{Wall~1}} \\[2pt]
Compound             & 399   &   2 & 55.33 & 0.13 &  0.43 \\
Trapped Lovers (2D)  & 5000  &   2 & 90.88 & 0.18 &  0.73 \\
Trapped Lovers (3D)  & 5000  &   3 & 90.88 & 0.18 &  0.73 \\
Isolation            & 9000  &   2 & 72.29 & 0.21 &  0.51 \\
Mk4                  & 1500  &   3 & 66.61 & 0.08 &  0.59 \\
\midrule
\multicolumn{6}{l}{\textit{Wall~2}} \\[2pt]
Iris                 & 150   &   4 & 86.78 & 0.85 &  0.02 \\
Mk3                  & 600   &   3 & 21.73 & 0.26 & $-$0.04 \\
G2mg\_128\_30        & 2048  & 128 & 25.69 & 0.70 & $-$0.44 \\
PBMC                 & 2638  &  50 & 81.64 & 0.62 &  0.19 \\
Digits               & 1797  &  64 & 50.94 & 0.10 &  0.41 \\
USPS                 & 9298  & 256 & 58.07 & 0.50 &  0.08 \\
HAR                  & 7352  & 561 & 64.24 & 0.58 &  0.06 \\
Fashion              & 35000 & 784 & 80.72 & 0.16 &  0.65 \\
\bottomrule
\end{tabular}
\caption{Optimal $q_z$ and $r$ parameters for SPORE across diagnostic exemplars. The final column computes $\frac{q_z}{100} - r$, representing the permissiveness of the Expansion phase of the optimal configuration per dataset.}
\label{tab:exemplar_properties}
\end{table}

The optimal configurations in Table~\ref{tab:exemplar_properties} suggest that Expansion and SCR form a continuum rather than two independent phases. Expansion permissiveness is jointly controlled by $q_z$ and $r$: larger $q_z$ values allow propagation into sparser regions, while larger $r$ values require stronger local support. The quantity $\frac{q_z}{100}-r$ therefore serves as a simple proxy for how aggressively Expansion is permitted to grow.

A clear trend emerges across the wall conditions. Wall~0 datasets consistently favor highly permissive Expansion ($0.43$--$0.83$), reflecting their strong geometric separation and limited boundary ambiguity. Wall~1 datasets occupy a similar but slightly more conservative regime ($0.43$--$0.73$), where the density-variance filter accommodates density heterogeneity while still allowing substantial direct recovery during Expansion. In contrast, several Wall~2 exemplars exhibit near-zero or negative permissiveness values, indicating that optimal performance is achieved by deliberately restricting Expansion. In these cases, Expansion isolates compact, high-confidence skeletons, while SCR performs a larger share of the final cluster recovery through density-normalized label propagation.

These results suggest that SPORE adapts its recovery strategy according to the dominant wall condition. As boundary ambiguity increases, the optimal operating point shifts away from direct expansion and toward a skeleton-first regime in which SCR assumes greater responsibility for constructing the final partition.

\FloatBarrier
\section{Conclusion}
This paper introduced SPORE, a clustering algorithm built around a separation of cluster discovery and boundary delineation. The central observation motivating its design is that density-based methods fail on two structurally distinct problems: variable-density recovery (Wall~1) and low-contrast boundary delineation (Wall~2), yet address both through a single mechanism, namely density connectivity. SPORE decouples these roles. An adaptive expansion phase grows cluster skeletons via per-cluster z-score thresholding, allowing each cluster to expand under its own density statistics rather than a shared global scale. A subsequent Small-Cluster Reassignment phase resolves ambiguous boundaries via $k$-NN majority voting, a signal that remains informative wherever neighborhood composition reflects cluster membership, regardless of whether a density gradient is present.

Evaluated across 28 benchmark datasets spanning 2D to 784D, SPORE achieves the highest mean ARI of any method tested, with statistically significant margins over all baselines at $p < 0.01$. Compared to density-based baselines, it occupies the Pareto-optimal region of the recovery--runtime tradeoff and reaches average $\text{ARI} \geq 0.75$ within five random-search trials, whereas no baseline attains this performance at any budget. The gains are consistent across both wall conditions and extend to standard topological benchmarks, demonstrating that improved robustness to density asymmetry and low-contrast boundaries does not come at the cost of classical density-based representational capacity.

Certain limitations remain. Relative to centroid-based methods such as K-means, there remains a gap in recovery capacity on convex clusters with low separation or high dimensionality. In this regime, K-means remains particularly effective because its iterative averaging and nearest-centroid assignment mechanism can amplify weak global separation signals even when $k$-NN structure is degraded by distance concentration. SPORE, by contrast, remains dependent on the existence of sufficiently reliable cluster cores from which skeletons can be recovered. Performance also declines on datasets near the intrinsic recovery limit, such as Fashion, where all methods plateau below 0.35 ARI, reflecting the fundamental difficulty of cluster separation in these settings.

More broadly, SPORE demonstrates that cluster discovery and boundary delineation are separable problems whose signals can be modeled independently to improve clustering performance. The results suggest that density connectivity is not the only viable mechanism for delineating clusters once they have been discovered, and that treating discovery and delineation as distinct tasks can substantially expand the effective recovery capacity of density-based clustering. Future work will formalize conditions for cluster recovery under the SPORE framework and explore methods for strengthening $k$-NN graph construction to improve skeleton recovery and propagation in high-dimensional settings.

\bibliographystyle{plainnat}
\bibliography{references}

\appendix

\section{Experiment Datasets}
\label{app:rep_cap_datasets}

Below is a description for each of the 28 datasets used in the Representational Capacity experiment.

\begin{enumerate}


\item \textbf{Compound (2D)}~\cite{gagolewski2022benchmark}: A synthetic dataset of five clusters of variable density, two of which are a dense cluster embedded within a sparse one.

\item \textbf{Spiral (2D)}~\cite{gagolewski2022benchmark}: A synthetic dataset consisting of three curved clusters that appear as a spiral shape.

\item \textbf{Flame (2D)}~\cite{gagolewski2022benchmark}: A two-cluster synthetic dataset composed of one compact group and one elongated, curved structure with partial overlap.

\item \textbf{Smile (2D)}~\cite{gagolewski2022benchmark}: A synthetic dataset composed of several non-convex components, including ring-like and curved structures arranged in a smile-like configuration.

\item \textbf{Wingnut (2D)}~\cite{gagolewski2022benchmark}: A synthetic dataset consisting of rectangular regions separated by a clear low-density gap. The left rectangle possesses a density peak in the upper right corner while the right rectangle possesses a density peak in the lower left corner.

\item \textbf{Trapped Lovers (2D)}: A 2D PCA-projection of the 3D variant, producing overlapping dense and sparse regions. 

\item \textbf{Aggregation (2D)}~\cite{gagolewski2022benchmark}: A collection of mostly well-separated globular clusters, some of which are weakly connected by thin bridging regions.

\item \textbf{Isolation (2D)}~\cite{gagolewski2022benchmark}: A synthetic dataset consisting of three concentric ring-shaped clusters with strongly varying densities.


\item \textbf{Trapped Lovers (3D)}~\cite{gagolewski2022benchmark}: A three-dimensional synthetic dataset consisting of two dense clusters embedded within a surrounding sparse structure. 

\item \textbf{Chainlink (3D)}~\cite{gagolewski2022benchmark}: Two interlocked toroidal clusters arranged orthogonally.

\item \textbf{Mk3 (3D)}~\cite{gagolewski2022benchmark}: A mixture of three Gaussian clusters, where two clusters are close and partially overlapping, and the third is well-separated.

\item \textbf{Mk4 (3D)}~\cite{gagolewski2022benchmark}: A synthetic configuration consisting of a dense central cluster together with two extended spiral-like structures arranged along a vertical axis.

\item \textbf{Tetra (3D)}~\cite{gagolewski2022benchmark}: Four Gaussian clusters arranged in a tetrahedral configuration, with moderate separation between clusters.

\item \textbf{Fish (3D)}~\cite{tawei2019fish}: A synthetic dataset of 9 fish species generated from length-weight statistics of \citet{hossain2015} for the Tetulia River, Bangladesh. Features are length, weight, and w/l ratio; the ratio feature induces high intra-cluster density variation and mild-to-moderate inter-cluster proximity.


\item \textbf{Iris (4D)}~\cite{Dua2019}: The classical Iris dataset consisting of measurements from three iris species, where one class is well-separated and the remaining two partially overlap.

\item \textbf{Banknote (4D)}~\cite{Dua2019}: A dataset derived from image features of genuine and forged banknotes, forming two moderately overlapping classes.

\item \textbf{Ecoli (7D)}~\cite{gagolewski2022benchmark}: A protein localization dataset with multiple classes and moderate class overlap.

\item \textbf{Seeds (7D)}~\cite{gagolewski2022benchmark}: Measurements of wheat kernels from three varieties, exhibiting overlapping feature distributions.

\item \textbf{Wine (13D)}~\cite{gagolewski2022benchmark}: Chemical analysis data from three wine cultivars, with moderate class separability.

\item \textbf{Pendigits (16D)}~\cite{Dua2019}: A handwritten digit dataset represented by pen-trajectory features, containing ten classes with significant intra-class variability.

\item \textbf{WDBC (30D)}~\cite{gagolewski2022benchmark}: The Wisconsin Diagnostic Breast Cancer dataset, consisting of two classes that are only weakly separable.

\item \textbf{PBMC\_3k (50D)}~\cite{Wolf2018Scanpy}: A single-cell RNA sequencing dataset containing multiple immune cell types. Reduced to 50D from the original 1838D for experimentation.

\item \textbf{Digits (64D)}~\cite{Pedregosa2011}: A handwritten digit dataset represented by pixel intensities, with ten classes and substantial class overlap.


\item \textbf{G2mg\_128\_20 (128D)}~\cite{gagolewski2022benchmark}: A synthetic two-Gaussian mixture with well-separated components in the original high-dimensional space.

\item \textbf{G2mg\_128\_30 (128D)}~\cite{gagolewski2022benchmark}: A synthetic two-Gaussian mixture with weaker separation, resulting in partial overlap between the components.


\item \textbf{USPS (256D)}~\cite{hull1994database}: A handwritten digit dataset with varying writing styles and class overlap.

\item \textbf{HAR\_Train\_Subset (561D)}~\cite{HARKaggle}: The training subset of a human activity recognition dataset derived from wearable sensor signals, containing multiple activity classes with overlapping feature distributions.

\item \textbf{Fashion (784D)}~\cite{gagolewski2022benchmark}: A Fashion-MNIST-based dataset consisting of grayscale clothing images from ten categories, characterized by high visual similarity and noise.

\end{enumerate}

\section{Hyperparameter Ranges}
\label{app:search_bounds}

\begin{table}[ht]
\centering
\caption{SPORE hyperparameter search bounds and fixed parameters.}
\begin{tabular}{lll}
\toprule
\textbf{Parameter} & \textbf{Meaning} & \textbf{Value / Search range} \\
\midrule
$q_z$ & Bounded reparameterization of $z$ (\S\ref{sec:proposed_method}) & $[12.5,\ 100]$ \\
\midrule
$r$   & Retention rate & $[0,\ 0.875]$ \\
\midrule
$M^*$ & \shortstack[l]{Cluster exponent;\\ internally $M = \lfloor N^{M^*} \rfloor$} & $[0.3,\ 0.7]$ \\
\midrule
$k_{\text{Expansion}}$ & Neighbor count for expansion step       & $[\lfloor \log_2 N \rfloor,\, 3\lfloor \log_2 N \rfloor]$ \\
\midrule
$q_{z_{SCR}}$ & Bounded reparameterization of $z_{SCR}$ (\S\ref{sec:proposed_method}) & $100$ \\
\midrule
$k_{\text{density}}$   & Neighbor count for density estimation  & $\lfloor \frac{k_{\text{Expansion}}}{2} \rfloor$ \\
\midrule
$k_{\text{SCR}}$       & Neighbor count for SCR                  & $\min(M,\ 2k_{\text{Expansion}})$ \\
\bottomrule
\end{tabular}
\end{table}

\begin{table}[ht]
\centering
\caption{DBSCAN hyperparameter search bounds.}
\begin{tabular}{lll}
\toprule
\textbf{Parameter} & \textbf{Meaning} & \textbf{Search range} \\
\midrule
\texttt{min\_samples} & \shortstack[l]{Minimum points for a core point;\\ scales with $D$~\citep{sander1998}} & $\left[\min(\sqrt{N}, D),\ \min(N, 3\sqrt{N}, 3D)\right]$ \\
\midrule
$q_\varepsilon$ & \shortstack[l]{$\varepsilon$ as the $q_\varepsilon$th percentile of\\ the $k$-NN distance distribution} & $[12.5,\ 100]$ \\
\bottomrule
\end{tabular}
\end{table}

\begin{table}[ht]
\centering
\caption{HDBSCAN hyperparameter search bounds.}
\begin{tabular}{lll}
\toprule
\textbf{Parameter} & \textbf{Meaning} & \textbf{Search range} \\
\midrule
\texttt{min\_cluster\_size} & Minimum cluster size & $\left[\lfloor N^{0.3} \rfloor,\ \lfloor N^{0.7} \rfloor\right]$ \\
\midrule
\texttt{min\_samples} & Controls mutual reachability smoothing & $[1,\ 2\lfloor \log_2 N \rfloor]$ \\
\midrule
\texttt{cluster\_selection\_method} & Cluster extraction strategy & $\{\texttt{leaf},\ \texttt{eom}\}$ \\
\bottomrule
\end{tabular}
\end{table}

\begin{table}[ht]
\centering
\caption{DPC hyperparameter search bounds.}
\begin{tabular}{lll}
\toprule
\textbf{Parameter} & \textbf{Meaning} & \textbf{Search range} \\
\midrule
$q_{dc}$ & \shortstack[l]{$d_c$ as the $q_{dc}$th percentile\\ of pairwise distances} & $[0,\ 50]$ \\
\midrule
$k$ & \shortstack[l]{Number of density peaks selected\\ as centers via $\gamma$-score ranking} & True $k$ (provided) \\
\bottomrule
\end{tabular}
\end{table}

\begin{table}[ht]
\centering
\caption{SNN-DBSCAN hyperparameter search bounds and fixed parameters.}
\begin{tabular}{lll}
\toprule
\textbf{Parameter} & \textbf{Meaning} & \textbf{Value / Search range} \\
\midrule
\texttt{min\_samples} & Minimum SNN density for a core point & $\left[\min(\sqrt{N}, D),\ \min(N, 3\sqrt{N}, 3D)\right]$ \\
\midrule
$f$ & \shortstack[l]{Shared-neighbor threshold as fraction of $k$, \\ analogous to SPORE's $r$; \\ computed as \texttt{eps} $= \lfloor f \cdot k \rfloor$ for DBSCAN } & $[0,\ 0.875]$ \\
\midrule
$k$ & \shortstack[l]{Shared-neighbor list length} & $[\lfloor \log_2 N \rfloor,\ \lfloor \sqrt{N} \rfloor]$ \\
\bottomrule
\end{tabular}
\end{table}

\begin{table}[ht]
\centering
\caption{K-means configuration.}
\begin{tabular}{lll}
\toprule
\textbf{Parameter} & \textbf{Meaning} & \textbf{Value} \\
\midrule
\texttt{n\_clusters} & Number of clusters & True $k$ (provided) \\
\midrule
\texttt{init}        & Initialization     & \texttt{k-means++} \\
\midrule
\texttt{n\_init}     & Number of restarts & $5$ \\
\bottomrule
\end{tabular}
\end{table}

\FloatBarrier
\section{Ablations}
\label{app:ablations}

To assess the contribution of key components of SPORE, the recovery-capacity experiment described in \S\ref{sec:rep_cap_experiment} is repeated with a different component of SPORE modified or removed each time. The following ablations are considered:
\begin{enumerate}
    \item \textbf{Nearest neighbor computation} (SPORE-ANN): approximate
    nearest neighbors replace exact $k$-NN, assessing whether the
    approximation introduces meaningful error relative to the exact variant.
    \item \textbf{SCR} (SPORE-MSR-0): \texttt{max\_rounds} is fixed to 0, effectively removing SCR.
    \item \textbf{Cluster seeding} (SPORE-RandSeed): cluster centers are
    initialized randomly rather than in density order, isolating the
    contribution of temporal shielding.
\end{enumerate}

\begin{figure}[ht]
    \centering
    \includegraphics[width=0.6\textwidth]{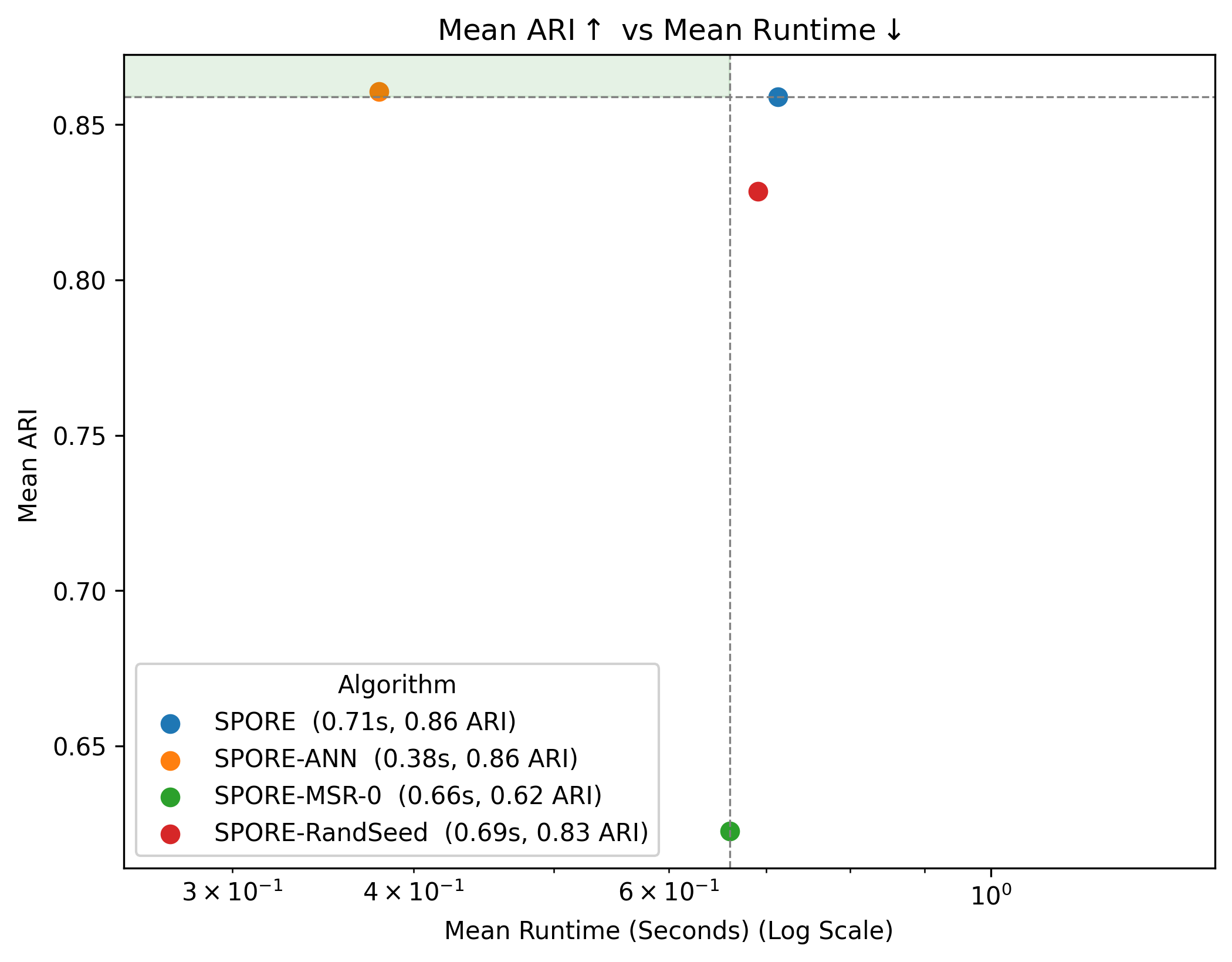}
    \caption{Mean ARI versus mean runtime (log scale) of ablated SPORE variants across 28 datasets. The shaded region corresponds to methods that achieve equivalent or higher ARI and equivalent or lower runtime than the second-best on each axis. SPORE-ANN, which achieves virtually identical ARI to SPORE at lower runtime, is the only variant within this region.}
    \label{fig:pareto-ablations}
\end{figure}

\FloatBarrier
\subsection{Approximate vs Exact Nearest Neighbors (SPORE-ANN)}

\begin{table}[ht]
\centering
\caption{Wilcoxon Signed Rank Test for ARI (SPORE-ANN vs.\ Others)}
\label{tab:wilcoxon_ann}
\begin{tabular}{lrrrrrc}
\hline
\textbf{Method} & \textbf{N\_pairs} & \textbf{Median Diff} & \textbf{Wilcoxon Stat} & \textbf{$p$-raw} & \textbf{$p$-corrected} & \textbf{Significant} \\
\hline
SPORE           & 28 & 0.00000 &  11.0 & 0.172616 & 0.172616 & $\times$ \\
SPORE-MSR-0     & 28 & 0.19095 & 276.0 & 0.000014 & 0.000041 & \checkmark \\
SPORE-RandSeed  & 28 & 0.00000 & 147.0 & 0.003699 & 0.007399 & \checkmark \\
\hline
\end{tabular}
\end{table}

Approximate neighbors introduce small perturbations into local distances, but the Expansion phase does not require exactly accurate distances. So long as the provided neighbors do not significantly inflate the estimated density statistics, the Expansion phase can function correctly. After repeating the representational capacity experiment (\S~\ref{sec:rep_cap_experiment}) with an HNSW~\cite{Malkov2018} for approximate nearest neighbors, the median difference in ARI between the approximate and exact variants was 0, and the difference was not significant under a Wilcoxon signed-rank test ($p = 0.17$).

On the other hand, runtime performance changed substantially. The ANN variant averaged
0.38\,s against the ENN variant's 0.71\,s, a reduction of approximately
46\%. This places SPORE-ANN in the Pareto-optimal position over all
methods evaluated, including the ENN variant, at the same mean ARI of
$\approx 0.86$.

\subsection{Removing SCR (SPORE-MSR-0)}
This ablation disables Small-Cluster Reassignment by fixing
\texttt{max\_rounds} to $0$, so that no post-expansion consolidation
occurs. Removing SCR produces a large drop in mean ARI (0.62 vs.\ 0.86),
with especially prominent losses on datasets with smooth transitions between
clusters (e.g., \textit{Mk3}:\ $0.51$, \textit{Wine}:\ $0.52$,
\textit{G2mg\_128\_30}:\ $0.01$). These datasets are often Gaussian-like
or convex, with high boundary proximity or overlap that creates smooth density
gradients between clusters. In such settings, expansion must remain
conservative to avoid under-segmentation. Without SCR to coalesce the resulting
fragments, the recovered clusters remain fragmented, substantially impairing
recovery.

\subsection{Random Cluster Seeding (SPORE-RandSeed)}
SPORE's default seeding order sorts candidate seeds by ascending $k$-NN
distance, causing expansion to initiate in high-confidence interior regions
before approaching ambiguous boundaries. This ordering induces temporal
shielding, in which dense regions discovered early are protected from being
absorbed into later expansions from sparser regions. It also causes early
expansions to estimate more conservative distance statistics, reducing the risk
of over-merging.

Random seeding removes these effects. Expansion may instead begin near boundary
regions or within higher-variance, lower-density regions, increasing the risk
that adjacent clusters are merged. This mechanism is consistent with the
observed losses on \texttt{Iris} ($0.89 \to 0.68$), \texttt{WDBC} ($0.78 \to
0.63$), \texttt{G2mg\_128\_30} ($0.89 \to 0.76$), and
\texttt{HAR\_Train\_Subset} ($0.60 \to 0.50$). These datasets all contain
ambiguous or overlapping boundaries; without density-ordered seeding, expansion
has an increased likelihood of initiating from unstable boundary regions and
merging adjacent clusters.

A noteworthy case is \texttt{Trapped Lovers} (2D variant) ($0.85 \to 0.78$),
where the random-seed variant exhibits reduced recovery without catastrophic
failure. This is likely because high-density regions remain structurally favored
even under random seed selection, since they contain more points within a local
geometric region. As a result, an approximate form of temporal shielding may
still arise empirically, though it is less reliable than explicitly seeding from
the highest-density candidate points.

\begin{table}[htbp]
\centering
\caption{ARI scores for SPORE ablation variants. Bold indicates the maximum per row. Standard deviations below 0.005 are omitted.}
\resizebox{\textwidth}{!}{
\begin{tabular}{lcccc}
\toprule
\textbf{Dataset} & \textbf{SPORE} & \textbf{SPORE-ANN} & \textbf{SPORE-MSR-0} & \textbf{SPORE-RandSeed} \\
\midrule
Compound            &           0.98 &           0.98 &           0.98 & \textbf{1.0} \\
Spiral              & \textbf{1.0}   & \textbf{1.0}   & \textbf{1.0}   & \textbf{1.0} \\
Flame               & \textbf{0.97}  & \textbf{0.97}  &           0.88 & \textbf{0.97} \\
Smile               & \textbf{1.0}   & \textbf{1.0}   & \textbf{1.0}   & \textbf{1.0} \\
Wingnut             & \textbf{1.0}   & \textbf{1.0}   & \textbf{1.0}   & \textbf{1.0} \\
Aggregation         & \textbf{0.99}  & \textbf{0.99}  &           0.91 & \textbf{0.99} \\
Isolation           & \textbf{1.0}   & \textbf{1.0}   & \textbf{1.0}   & \textbf{1.0} \\
Trapped Lovers (2D) & \textbf{0.85}  & \textbf{0.85}  & \textbf{0.85}  &           0.78 \\
Trapped Lovers (3D) & \textbf{1.0}   & \textbf{1.0}   &           0.99 & \textbf{1.0} \\
Chainlink           & \textbf{1.0}   & \textbf{1.0}   & \textbf{1.0}   & \textbf{1.0} \\
Mk3                 & \textbf{0.85}  & \textbf{0.85}  &           0.51 &           0.83 \\
Mk4                 & \textbf{1.0}   & \textbf{1.0}   & \textbf{1.0}   & \textbf{1.0} \\
Tetra               & \textbf{1.0}   & \textbf{1.0}   &           0.94 &           0.94 \\
Fish                & \textbf{0.60}  & \textbf{0.60}  &           0.25 &           0.59 \\
Iris                & \textbf{0.89}  & \textbf{0.89}  &           0.59 &           0.68 \\
Banknote            & \textbf{0.96}  & \textbf{0.96}  &           0.68 & \textbf{0.96} \\
Ecoli               &           0.73 &           0.73 &           0.49 & \textbf{0.74} \\
Seeds               & \textbf{0.77}  & \textbf{0.77}  &           0.46 & \textbf{0.77} \\
Wine                &           0.90 &           0.90 &           0.52 & \textbf{0.92} \\
Pendigits           & \textbf{0.76}  & \textbf{0.76}  &           0.63 & \textbf{0.76} \\
WDBC                & \textbf{0.78}  & \textbf{0.78}  &           0.31 &           0.63 \\
PBMC\_3k            & \textbf{0.81}  & \textbf{0.81}  &           0.13 &           0.80 \\
Digits              & \textbf{0.85}  & \textbf{0.85}  &           0.48 &           0.79 \\
G2mg\_128\_20       & \textbf{1.0}   & \textbf{1.0}   &           0.04 & \textbf{1.0} \\
G2mg\_128\_30       & \textbf{0.89}  & \textbf{0.89}  &           0.01 &           0.76 \\
USPS                & \textbf{0.64}  & \textbf{0.64}  &           0.36 &           0.62 \\
HAR\_Train\_Subset  & \textbf{0.60}  & \textbf{0.60}  &           0.27 &           0.50 \\
Fashion             &           0.23 & \textbf{0.28}  &           0.15 &           0.17 \\
\bottomrule
\end{tabular}
}
\label{tab:ablation_ari}
\end{table}

\FloatBarrier
\section{Representational Capacity Results}
\label{app:full_repr_cap_results}

\begin{table}[ht]
\centering
\caption{ARI scores across datasets and methods. Bold indicates the maximum value per row. Standard deviations below 0.005 are omitted.}
\label{tab:full_ari}
\resizebox{\textwidth}{!}{%
\begin{tabular}{lcccccc}
\toprule
Dataset $(N, D, D')$ & DBSCAN & DPC & HDBSCAN & KMeans & SNN\_DBSCAN & SPORE \\
\midrule
Compound (399, 2, 2)               & 0.87          & 0.82          & 0.84          & 0.59 $\pm$ 0.01          & 0.86          & \textbf{0.98} \\
Spiral (312, 2, 2)                 & \textbf{1.0}  & \textbf{1.0}  & \textbf{1.0}  & $-$0.01                  & \textbf{1.0}  & \textbf{1.0}  \\
Flame (240, 2, 2)                  & 0.91          & \textbf{1.0}  & 0.92          & 0.46 $\pm$ 0.02          & 0.95          & 0.97          \\
Smile (1000, 2, 2)                 & 0.97          & 0.95          & \textbf{1.0}  & 0.48 $\pm$ 0.06          & 0.99          & \textbf{1.0}  \\
Wingnut (1016, 2, 2)               & 0.99          & \textbf{1.0}  & \textbf{1.0}  & 0.44                     & 0.97          & \textbf{1.0}  \\
Aggregation (788, 2, 2)            & 0.97          & \textbf{1.0}  & 0.91          & 0.73 $\pm$ 0.01          & 0.90          & 0.99          \\
Isolation (9000, 2, 2)             & 0.99          & 0.02          & \textbf{1.0}  & $-$0.0                   & \textbf{1.0}  & \textbf{1.0}  \\
Trapped Lovers (5000, 3, 2)        & 0.53          & 0.32          & 0.64          & 0.15                     & 0.0           & \textbf{0.85} \\
Trapped Lovers (5000, 3, 3)        & \textbf{1.0}  & 0.49          & \textbf{1.0}  & 0.15                     & \textbf{1.0}  & \textbf{1.0}  \\
Chainlink (1000, 3, 3)             & \textbf{1.0}  & \textbf{1.0}  & \textbf{1.0}  & 0.09                     & \textbf{1.0}  & \textbf{1.0}  \\
Mk3 (600, 3, 3)                    & 0.54          & 0.87          & 0.57          & \textbf{0.88}            & 0.57          & 0.85          \\
Mk4 (1500, 3, 3)                   & 0.61          & 0.76          & \textbf{1.0}  & 0.41                     & \textbf{1.0}  & \textbf{1.0}  \\
Tetra (400, 3, 3)                  & 0.88          & \textbf{1.0}  & 0.98          & \textbf{1.0}             & 0.98          & \textbf{1.0}  \\
Fish (4080, 3, 3)                  & \textbf{0.87} & \textbf{0.87} & \textbf{0.87} & 0.81                     & 0.78          & 0.60          \\
Iris (150, 4, 4)                   & 0.62          & \textbf{0.89} & 0.57          & 0.72                     & 0.88          & \textbf{0.89} \\
Banknote (1372, 4, 4)              & 0.80          & \textbf{0.97} & 0.82          & 0.02                     & 0.56          & 0.96          \\
Ecoli (336, 7, 7)                  & 0.52          & 0.48          & 0.41          & 0.43 $\pm$ 0.01          & 0.71          & \textbf{0.73} \\
Seeds (210, 7, 7)                  & 0.53          & 0.76          & 0.42          & 0.70                     & \textbf{0.82} & 0.77          \\
Wine (178, 13, 13)                 & 0.46          & 0.80          & 0.53          & 0.85 $\pm$ 0.01          & 0.83          & \textbf{0.90} \\
Pendigits (10992, 16, 16)          & 0.55          & 0.66          & 0.63          & 0.56 $\pm$ 0.03          & 0.58          & \textbf{0.76} \\
WDBC (569, 30, 30)                 & 0.46          & 0.45          & 0.23          & 0.71                     & 0.37          & \textbf{0.78} \\
PBMC\_3k (2638, 1838, 50)          & 0.27          & 0.35          & 0.05          & 0.75 $\pm$ 0.11          & 0.61          & \textbf{0.81} \\
Digits (1797, 64, 64)              & 0.41          & 0.54          & 0.65          & 0.66 $\pm$ 0.01          & 0.78          & \textbf{0.85} \\
G2mg\_128\_20 (2048, 128, 128)     & 0.0           & 0.86          & 0.06          & \textbf{1.0}             & 0.11          & \textbf{1.0}  \\
G2mg\_128\_30 (2048, 128, 128)     & 0.0           & 0.11          & 0.01          & \textbf{0.95}            & 0.0           & 0.89          \\
USPS (9298, 256, 256)              & 0.15          & 0.25          & 0.10          & 0.54 $\pm$ 0.02          & 0.36          & \textbf{0.64} \\
HAR\_Train\_Subset (7352, 561, 561)& 0.32          & 0.46          & 0.29          & 0.46                     & 0.28          & \textbf{0.60} \\
Fashion (35000, 784, 784)          & 0.05          & 0.13          & 0.02          & \textbf{0.35} $\pm$ 0.01 & 0.0           & 0.23          \\
\bottomrule
\end{tabular}%
}
\end{table}

\begin{table}[ht]
\centering
\caption{NMI scores across datasets and methods. Bold indicates the maximum value per row. Standard deviations below 0.005 are omitted.}
\label{tab:full_nmi}
\resizebox{\textwidth}{!}{%
\begin{tabular}{lcccccc}
\toprule
Dataset $(N, D, D')$ & DBSCAN & DPC & HDBSCAN & KMeans & SNN\_DBSCAN & SPORE \\
\midrule
Compound (399, 2, 2)               & 0.89          & 0.84          & 0.86          & 0.71 $\pm$ 0.01          & 0.90          & \textbf{0.97} \\
Spiral (312, 2, 2)                 & 0.99          & \textbf{1.0}  & \textbf{1.0}  & 0.0                      & \textbf{1.0}  & \textbf{1.0}  \\
Flame (240, 2, 2)                  & 0.85          & \textbf{1.0}  & 0.86          & 0.41 $\pm$ 0.03          & 0.91          & 0.93          \\
Smile (1000, 2, 2)                 & 0.92          & 0.94          & \textbf{1.0}  & 0.76 $\pm$ 0.02          & 0.98          & \textbf{1.0}  \\
Wingnut (1016, 2, 2)               & 0.97          & \textbf{1.0}  & \textbf{1.0}  & 0.34                     & 0.94          & \textbf{1.0}  \\
Aggregation (788, 2, 2)            & 0.96          & \textbf{0.99} & 0.95          & 0.84                     & 0.94          & \textbf{0.99} \\
Isolation (9000, 2, 2)             & 0.99          & 0.03          & \textbf{1.0}  & 0.0                      & \textbf{1.0}  & \textbf{1.0}  \\
Trapped Lovers (5000, 3, 2)        & 0.44          & 0.41          & 0.69          & 0.38                     & 0.0           & \textbf{0.82} \\
Trapped Lovers (5000, 3, 3)        & 0.99          & 0.67          & \textbf{1.0}  & 0.38                     & \textbf{1.0}  & \textbf{1.0}  \\
Chainlink (1000, 3, 3)             & \textbf{1.0}  & \textbf{1.0}  & \textbf{1.0}  & 0.07                     & \textbf{1.0}  & \textbf{1.0}  \\
Mk3 (600, 3, 3)                    & 0.66          & 0.84          & 0.72          & \textbf{0.85}            & 0.72          & 0.81          \\
Mk4 (1500, 3, 3)                   & 0.68          & 0.76          & 0.99          & 0.52                     & \textbf{1.0}  & 0.99          \\
Tetra (400, 3, 3)                  & 0.87          & \textbf{1.0}  & 0.97          & \textbf{1.0}             & 0.97          & \textbf{1.0}  \\
Fish (4080, 3, 3)                  & 0.95          & \textbf{0.96} & 0.95          & 0.91                     & 0.90          & 0.81          \\
Iris (150, 4, 4)                   & 0.66          & 0.86          & 0.73          & 0.74                     & 0.85          & \textbf{0.89} \\
Banknote (1372, 4, 4)              & 0.70          & \textbf{0.94} & 0.74          & 0.02                     & 0.63          & 0.93          \\
Ecoli (336, 7, 7)                  & 0.51          & 0.46          & 0.41          & 0.60 $\pm$ 0.01          & 0.67          & \textbf{0.69} \\
Seeds (210, 7, 7)                  & 0.59          & 0.72          & 0.49          & 0.67                     & \textbf{0.77} & 0.73          \\
Wine (178, 13, 13)                 & 0.57          & 0.80          & 0.62          & 0.84 $\pm$ 0.02          & 0.82          & \textbf{0.88} \\
Pendigits (10992, 16, 16)          & 0.72          & 0.78          & 0.76          & 0.69 $\pm$ 0.01          & 0.78          & \textbf{0.83} \\
WDBC (569, 30, 30)                 & 0.34          & 0.44          & 0.26          & 0.60                     & 0.46          & \textbf{0.65} \\
PBMC\_3k (2638, 1838, 50)          & 0.27          & 0.52          & 0.17          & \textbf{0.83} $\pm$ 0.03 & 0.69          & 0.82          \\
Digits (1797, 64, 64)              & 0.69          & 0.74          & 0.80          & 0.74                     & 0.83          & \textbf{0.89} \\
G2mg\_128\_20 (2048, 128, 128)     & 0.0           & 0.78          & 0.23          & \textbf{1.0}             & 0.27          & \textbf{1.0}  \\
G2mg\_128\_30 (2048, 128, 128)     & 0.0           & 0.08          & 0.07          & \textbf{0.90}            & 0.04          & 0.82          \\
USPS (9298, 256, 256)              & 0.33          & 0.45          & 0.42          & 0.63 $\pm$ 0.01          & 0.54          & \textbf{0.72} \\
HAR\_Train\_Subset (7352, 561, 561)& 0.47          & 0.66          & 0.43          & 0.59                     & 0.47          & \textbf{0.71} \\
Fashion (35000, 784, 784)          & 0.11          & 0.37          & 0.07          & \textbf{0.52}            & 0.02          & 0.45          \\
\bottomrule
\end{tabular}%
}
\end{table}

\begin{table}[ht]
\centering
\caption{Runtime (seconds) across datasets and methods. Bold indicates the minimum value per row. Standard deviations below 0.005 are omitted.}
\label{tab:runtime}
\resizebox{\textwidth}{!}{%
\begin{tabular}{lcccccc}
\toprule
Dataset $(N, D, D')$ & DBSCAN & DPC & HDBSCAN & KMeans & SNN\_DBSCAN & SPORE \\
\midrule
Compound (399, 2, 2)               & \textbf{0.0}          & \textbf{0.0}          & \textbf{0.0}          & 0.18 $\pm$ 0.42          & 0.05                  & 0.03                  \\
Spiral (312, 2, 2)                 & \textbf{0.0}          & \textbf{0.0}          & \textbf{0.0}          & 0.18 $\pm$ 0.42          & 0.02                  & 0.02                  \\
Flame (240, 2, 2)                  & \textbf{0.0}          & \textbf{0.0}          & \textbf{0.0}          & 0.03 $\pm$ 0.01          & 0.01                  & 0.02                  \\
Smile (1000, 2, 2)                 & \textbf{0.0}          & 0.02                  & 0.01                  & 0.05 $\pm$ 0.01          & 0.37 $\pm$ 0.01       & 0.04 $\pm$ 0.01       \\
Wingnut (1016, 2, 2)               & \textbf{0.0}          & 0.02                  & 0.01                  & 0.04 $\pm$ 0.01          & 0.39 $\pm$ 0.01       & 0.04                  \\
Aggregation (788, 2, 2)            & \textbf{0.0}          & 0.01                  & 0.01                  & 0.05 $\pm$ 0.01          & 0.16 $\pm$ 0.01       & 0.03                  \\
Isolation (9000, 2, 2)             & \textbf{0.04}         & 1.63 $\pm$ 0.03       & 0.09 $\pm$ 0.01       & \textbf{0.04}            & 35.59 $\pm$ 0.22      & 0.16 $\pm$ 0.01       \\
Trapped Lovers (5000, 3, 2)        & \textbf{0.02}         & 0.47 $\pm$ 0.01       & 0.06                  & 0.04 $\pm$ 0.01          & 5.63 $\pm$ 0.04       & 0.10 $\pm$ 0.01       \\
Trapped Lovers (5000, 3, 3)        & \textbf{0.04}         & 0.49 $\pm$ 0.01       & 0.08                  & 0.05 $\pm$ 0.01          & 6.71 $\pm$ 0.04       & 0.12 $\pm$ 0.01       \\
Chainlink (1000, 3, 3)             & \textbf{0.01}         & 0.02                  & \textbf{0.01}         & 0.05 $\pm$ 0.01          & 0.23 $\pm$ 0.01       & 0.03                  \\
Mk3 (600, 3, 3)                    & \textbf{0.0}          & 0.01                  & 0.01                  & 0.04 $\pm$ 0.01          & 0.15 $\pm$ 0.01       & 0.03                  \\
Mk4 (1500, 3, 3)                   & \textbf{0.01}         & 0.04                  & 0.02                  & 0.05 $\pm$ 0.01          & 0.59 $\pm$ 0.02       & 0.05 $\pm$ 0.01       \\
Tetra (400, 3, 3)                  & \textbf{0.0}          & \textbf{0.0}          & 0.01                  & 0.04 $\pm$ 0.01          & 0.04                  & 0.03                  \\
Fish (4080, 3, 3)                  & \textbf{0.03}         & 0.29 $\pm$ 0.01       & 0.05                  & \textbf{0.03} $\pm$ 0.01 & 6.87 $\pm$ 0.06       & 0.13 $\pm$ 0.01       \\
Iris (150, 4, 4)                   & \textbf{0.0}          & \textbf{0.0}          & \textbf{0.0}          & 0.03                     & 0.01                  & 0.02                  \\
Banknote (1372, 4, 4)              & \textbf{0.01}         & 0.04                  & 0.02                  & 0.05                     & 0.90 $\pm$ 0.01       & 0.07 $\pm$ 0.02       \\
Ecoli (336, 7, 7)                  & \textbf{0.0}          & \textbf{0.0}          & 0.01                  & 0.04 $\pm$ 0.01          & 0.03                  & 0.02                  \\
Seeds (210, 7, 7)                  & \textbf{0.0}          & \textbf{0.0}          & \textbf{0.0}          & 0.03 $\pm$ 0.01          & 0.01                  & 0.02                  \\
Wine (178, 13, 13)                 & \textbf{0.0}          & \textbf{0.0}          & \textbf{0.0}          & 0.03                     & 0.01                  & 0.02                  \\
Pendigits (10992, 16, 16)          & 0.57 $\pm$ 0.41       & 2.80 $\pm$ 0.03       & 1.51 $\pm$ 0.02       & \textbf{0.09} $\pm$ 0.01 & 27.78 $\pm$ 0.45      & 0.64 $\pm$ 0.43       \\
WDBC (569, 30, 30)                 & 0.04                  & \textbf{0.01}         & \textbf{0.01}         & 0.04                     & 0.19 $\pm$ 0.01       & 0.05 $\pm$ 0.01       \\
PBMC\_3k (2638, 1838, 50)          & 0.17 $\pm$ 0.01       & 0.17                  & 0.19                  & \textbf{0.08} $\pm$ 0.02 & 3.36 $\pm$ 0.03       & 0.24 $\pm$ 0.01       \\
Digits (1797, 64, 64)              & \textbf{0.07}         & 0.09                  & 0.15 $\pm$ 0.01       & \textbf{0.07} $\pm$ 0.01 & 1.53 $\pm$ 0.07       & 0.17 $\pm$ 0.02       \\
G2mg\_128\_20 (2048, 128, 128)     & 0.11 $\pm$ 0.01       & 0.36 $\pm$ 0.01       & 0.38                  & \textbf{0.05} $\pm$ 0.01 & 2.71 $\pm$ 0.03       & 0.23 $\pm$ 0.01       \\
G2mg\_128\_30 (2048, 128, 128)     & 0.11 $\pm$ 0.01       & 0.34                  & 0.39 $\pm$ 0.01       & \textbf{0.06} $\pm$ 0.01 & 2.75 $\pm$ 0.02       & 0.22 $\pm$ 0.01       \\
USPS (9298, 256, 256)              & 1.38 $\pm$ 0.02       & 10.28 $\pm$ 0.18      & 16.71 $\pm$ 0.09      & \textbf{0.33} $\pm$ 0.03 & 20.00 $\pm$ 0.24      & 0.90 $\pm$ 0.02       \\
HAR\_Train\_Subset (7352, 561, 561)& 1.53 $\pm$ 0.02       & 18.15 $\pm$ 0.25      & 28.55 $\pm$ 0.65      & \textbf{0.36} $\pm$ 0.02 & 12.58 $\pm$ 0.11      & 1.11 $\pm$ 0.02       \\
Fashion (35000, 784, 784)          & 23.64 $\pm$ 0.11      & 798.41                & 892.09                & \textbf{4.89} $\pm$ 0.56 & 345.18                & 15.44 $\pm$ 0.14      \\
\bottomrule
\end{tabular}%
}
\end{table}

\end{document}